\documentclass[11pt,letterpaper]{article}
\usepackage[margin=0.75in]{geometry}
\usepackage[T1]{fontenc}
\usepackage{newtxtext}
\usepackage{helvet}
\usepackage{courier}
\usepackage[hyphens]{url}
\usepackage{graphicx}
\usepackage[authoryear]{natbib}
\setcitestyle{aysep={}}
\usepackage{caption}
\usepackage{booktabs}
\usepackage{array}
\usepackage{tabularx}
\usepackage{placeins}
\usepackage{algorithm}
\usepackage{algorithmic}
\renewcommand{\algorithmicinput}{\textbf{Input:}}
\renewcommand{\algorithmicoutput}{\textbf{Output:}}
\renewcommand{\algorithmicfor}{for}
\renewcommand{\algorithmicforall}{for all}
\renewcommand{\algorithmicdo}{do}
\renewcommand{\algorithmicend}{end}
\usepackage{amsmath}
\usepackage{amssymb}
\usepackage{mathtools}
\usepackage{amsthm}
\usepackage{amsfonts}
\usepackage{microtype}
\usepackage[hidelinks]{hyperref}
\let\arxivfigure\figure
\let\endarxivfigure\endfigure
\renewenvironment{figure}[1][tbp]
  {\arxivfigure[!tbp]\centering\begin{minipage}{0.68\linewidth}}
  {\end{minipage}\endarxivfigure}
\expandafter\let\expandafter\arxivwidetable\csname table*\endcsname
\expandafter\let\expandafter\endarxivwidetable\csname endtable*\endcsname
\renewenvironment{table*}[1][tbp]
  {\arxivwidetable[#1]\renewcommand{\small}{\fontsize{9}{11}\selectfont}}
  {\endarxivwidetable}


\theoremstyle{plain}
\newtheorem{theorem}{Theorem}[section]
\newtheorem{proposition}[theorem]{Proposition}
\newtheorem{lemma}[theorem]{Lemma}
\newtheorem{corollary}[theorem]{Corollary}
\theoremstyle{definition}

\newtheorem{assumption}[theorem]{Assumption}
\theoremstyle{remark}

\DeclareMathOperator*{\argmin}{argmin}

\DeclareMathOperator{\Diag}{Diag}
\DeclareMathOperator{\diag}{diag}

\title{What Moved the Alarm Score?\\
Causal Attribution with Exact Computation\\
and Scarce-Incident Estimation}
\author{%
Masaaki Takada\\
Toshiba Corporation\\
\small\texttt{masaaki.takada.f65@mail.toshiba}
\and
Ryusei Shingaki\\
Toshiba Corporation\\
\small\texttt{ryusei.shingaki.v85@mail.toshiba}}
\date{}
\hypersetup{
  pdftitle={What Moved the Alarm Score? Causal Attribution with Exact Computation and Scarce-Incident Estimation},
  pdfauthor={Masaaki Takada and Ryusei Shingaki}
}

\begin{document}
\maketitle

\begin{abstract}
Alarms in industrial systems and scientific instruments often demand immediate diagnosis despite scarce incident data.
To guide intervention, an explanation must trace how changes in the data-generating process propagate through the causal structure to alter the alarm score, rather than merely flagging shifts that may be downstream effects.
Yet exact attribution across interacting causal components appears combinatorial, while existing methods often either assume an unchanged causal model or require enough incident data to estimate the changed system reliably.
We address this setting by formulating the alarm-score change as a Shapley game over hybrid structural causal model (SCM) states.
Treating structural functions and exogenous laws as distinct players yields a finer-grained, function/law-resolved report of how each type of change propagates to the alarm score.
The same framework defines realized-event and population reports, with an event--population coherence identity linking their attributions.
For ordered linear SCMs and quadratic scores, a causal first- and second-moment recursion combined with quadrature computes the full Shapley vector exactly in polynomial time, eliminating combinatorial computation.
To estimate this target from scarce incident data, we adapt a well-observed reference SCM to the incident regime using cross-fitted estimates of mechanism changes and residual moments, and derive attribution-error bounds and top-\(k\) recovery guarantees.
Experiments on controlled and measured shifts illustrate the diagnoses and assess scaling to hundreds of variables and estimation from small incident samples.
\end{abstract}

\section{Introduction}
\label{sec:introduction}

Alarms in industrial and scientific systems demand decisions before their
causes are clear.  Immediately after an alarm, incident data are scarce and
causal mechanisms may have changed.
Ranking variables by observed shift is unreliable: a large shift may be
downstream or have little effect on the alarm score, whereas a modest upstream
shift may propagate through the system and strongly affect the score.

The diagnostic question is therefore how changes in the data-generating
process propagated through the system to move a supplied alarm score, rather
than merely which variables shifted.  We seek an additive allocation of this
contrast to structural-function and exogenous-law changes.  Producing it soon
enough to guide intervention requires three coupled capabilities.  First, the
report must accommodate a changing SCM and resolve parent-dependent mechanisms
from exogenous inputs; otherwise opposing effects at one node can cancel and
obscure which kind of repair could help.  Second, to compute the report quickly,
its Shapley allocation must avoid factorial permutation averaging and
exponential coalition enumeration.  Third, to work from few incident samples,
it must leverage the reference regime before enough incident data accumulate
to relearn the SCM.  Resolution, computation, and estimation thus jointly
determine whether attribution can guide intervention immediately after an alarm.

Existing methods solve only subsets of this immediate-diagnosis problem.
Feature-attribution methods explain a fixed score map through observed inputs \citep{lundberg2017unified}; event-level causal attribution explains an individual prediction or outlier within one model \citep{heskes2020causal,budhathoki2022causal,nguyen2024noisy}; node-wise population attribution replaces a whole causal conditional \citep{budhathoki2021why,zhang2023model,quintasmartinez2024multiply}; and shift discovery locates changed components \citep{huang2020causal,chen2023iscan,seo2026fans}.
The first two hold generating mechanisms fixed.  Node-wise population reports
compare regimes, but they can require substantial target-regime data to
estimate conditional changes or hybrid values and can
merge opposing function/law effects.  Shift discovery stops before allocating
a supplied score contrast after propagation.
Locating where the system changed and determining what moved the alarm score
are therefore distinct tasks; none of these approaches provides the resolved,
tractable, and scarce-data attribution needed here.

Our starting point is a unified Shapley game over hybrid SCM states, indexed by
source, target, player granularity, and score.  Each coalition takes selected
components from the incident regime and the rest from the reference regime,
then evaluates the score after causal propagation.  The primary population
report uses structural functions and exogenous laws as separate players, so
their propagated contributions remain visible.  The same framework defines
realized-event reports, with an event--population
coherence identity linking their expected attributions to the population
report.  Because the allocation is relative to aligned endpoint SCMs, physical
interpretation still requires system-specific semantics.

To make this estimand available in the short incident window, we specialize to
ordered linear SCMs and quadratic scores.  Every hybrid expectation then
depends only on causally propagated first and second moments.
This includes common PCA- and ICA-based monitoring scores, quadratic scores formed from Mahalanobis or graphical-lasso precision matrices, and squared \(z\)-scores \citep{qin2003spm,lee2004ica_spm,mahalanobis1936distance,friedman2008glasso}.
This second-order closure resolves both remaining bottlenecks: a causal-order
moment recursion, shared backward-sensitivity sweep, and exact quadrature
compute the full Shapley vector exactly up to roundoff in polynomial time,
while a well-observed reference SCM anchors cross-fitted estimates of incident
mechanism deviations and residual moments.  The resulting player-specific
error bounds yield a uniform rate, sign recovery, and top-\(k\) recovery under
stated conditions.  Thus the same specialization makes the resolved report
both exactly computable and estimable from scarce incident data.

The resulting method is \textbf{S$^{4}$-RCA}: \textbf{S}ample-efficient, \textbf{S}CM-grounded, \textbf{S}hapley-consistent, and \textbf{S}hift-aware Root Cause Analysis.
Our contributions are:
\begin{itemize}
  \item \textbf{Function/law-resolved causal score attribution.} We introduce a source--target--player--score framework whose primary population game separates structural functions from exogenous laws and connects to realized-event attribution through event--population coherence.
  \item \textbf{Exact polynomial-time computation.} The moment program computes the full function/law vector, up to roundoff, in \(O(d^4)\) time and \(O(d^2)\) memory without enumerating orders or coalitions.
  \item \textbf{Scarce-incident estimation.} Reference-anchored cross-fitting yields player-specific error bounds and guarantees for sign and top-\(k\) recovery. Controlled and measured shifts assess resolution, scaling, recovery, and model-relative diagnosis.
\end{itemize}

\section{Unified Causal Score Attribution}
\label{sec:prelim}

\paragraph{Source, target, players, and score.}
\label{subsec:residual-setup}
The primary attribution problem has four objects:
(i) a normal-regime \emph{source world}, represented by an SCM \(M^{(0)}\);
(ii) an incident-regime \emph{target world}, represented by \(M^{(1)}\);
(iii) a set \(\mathcal U\) of SCM-state players specifying which changes are attributed jointly or separately; and
(iv) a supplied anomaly score \(\tau\), inducing a scalar functional \(T\) to decompose.
Together, \((M^{(0)},M^{(1)},\mathcal U,T)\) determine the source-to-target contrast, its decomposition granularity, and the causal estimand.

Let \(\mathcal X\subseteq\mathbb R^d\) be the observed state space, with
\(x=(x_j)_{j=1}^d\in\mathcal X\) and exogenous vector
\(e=(e_j)_{j=1}^d\).
For regime \(r\in\{0,1\}\), write
\(F^{(r)}=(F_j^{(r)})_{j=1}^d\),
\(Q^{(r)}=\bigotimes_{j=1}^d Q_j^{(r)}\), and
\(M^{(r)}=M(F^{(r)},Q^{(r)})\) for the ordered SCM
\[
x_j=F_j^{(r)}(x_{<j},e_j),\qquad e_j\sim Q_j^{(r)}.
\]
The exogenous coordinates are mutually independent, and
\(x=G_{F^{(r)}}(e)\) denotes the reduced form.
For the supplied anomaly score \(\tau:\mathcal X\to\mathbb R\), define
\begin{align}
T(M(F,Q)):=\mathbb E_{e\sim Q}[\tau(G_F(e))].
\label{eq:score-functional}
\end{align}
The primary population question attributes \(T(M^{(1)})-T(M^{(0)})\).
Source, target, and hybrid states must preserve variable and exogenous-coordinate meanings and have finite target values.

\paragraph{SCM-state players to attribute.}
\label{def:scm-state-replacement-game}
Let
\(
\mathcal A=\{\mathtt F_j,\mathtt Q_j:j\in[d]\}
\)
be the atomic switch labels.
Here \(\mathtt F_j\) labels the switch from the actual function \(F_j^{(0)}\) to \(F_j^{(1)}\), and \(\mathtt Q_j\) labels the switch from \(Q_j^{(0)}\) to \(Q_j^{(1)}\); the monospaced symbols are labels, not functions or laws themselves.
A player \(u\) is a nonempty subset of \(\mathcal A\) whose switches act jointly, and a player set \(\mathcal U\) is a collection of pairwise disjoint players.
A player \(u\) is \emph{node-local} if \(u\subseteq\{\mathtt F_j,\mathtt Q_j\}\)
for some \(j\); a player set is node-local if every player is.
The primary player set to attribute is \(\mathcal U=\mathcal U_{\mathtt F\cup\mathtt Q}\), with
\[
\mathcal U_{\mathtt F\cup\mathtt Q}
=\{\{\mathtt F_j\}:j\in[d]\}\cup\{\{\mathtt Q_j\}:j\in[d]\}.
\]
Its \(2d\) singleton players yield function/law-resolved attribution.
Other useful player sets are
\(\mathcal U_{(\mathtt F,\mathtt Q)}
=\{\{\mathtt F_j,\mathtt Q_j\}:j\in[d]\}\) for paired nodes and
\(\mathcal U_{\mathtt Q}=\{\{\mathtt Q_j\}:j\in[d]\}\).

A coalition is \(U\in2^{\mathcal U}\), equivalently \(U\subseteq\mathcal U\); thus \(u\in\mathcal U\) denotes one player, whereas \(U\) denotes a set of players.
The hybrid \(M^U=M(F^U,Q^U)\) takes selected components from \(M^{(1)}\) and the rest from \(M^{(0)}\):
\[
F_j^U=F_j^{\mathbf 1\{\mathtt F_j\in\bigcup_{w\in U}w\}},
\qquad
Q_j^U=Q_j^{\mathbf 1\{\mathtt Q_j\in\bigcup_{w\in U}w\}}.
\]
Each coalition is therefore an SCM replacement followed by causal propagation, not feature masking.
Switching the label pair \(\{\mathtt F_j,\mathtt Q_j\}\) jointly recovers the established node-wise population granularity; separating them answers whether the node's score effect arose through its parent-dependent function, its exogenous law, or both.

\paragraph{Causal Shapley estimand.}
\label{def:diagnostic-reports}
The causal question is how much each prespecified SCM-state player contributes
to the score contrast between the source and target worlds.  Each coalition
\(U\) answers the corresponding counterfactual subquestion by switching exactly
those players, inducing the finite game \(v:2^{\mathcal U}\to\mathbb R\),
\(v(U)=T(M^U)\).  We define player \(u\)'s causal estimand as its Shapley value
in this game, which averages its marginal expected-score change over all
component orders \citep{shapley1953value}.  Denote it by
\(\Phi_u(M^{(0)},M^{(1)},\mathcal U,T)\), abbreviated as \(\Phi_u\) when the
defining tuple is fixed:
\begin{align}
\Phi_u
&:=
\sum_{U\subseteq \mathcal U\setminus\{u\}}
\frac{|U|!(|\mathcal U|-|U|-1)!}{|\mathcal U|!}
\notag\\
&\quad\times
\left\{T(M^{U\cup\{u\}})-T(M^U)\right\}.
\label{eq:causal-shapley}
\end{align}
For singleton players, we suppress braces in attribution subscripts, writing \(\Phi_{\mathtt F_j}:=\Phi_{\{\mathtt F_j\}}\) and similarly for \(\mathtt Q_j\).
For the primary population shift, take \(\mathcal U=\mathcal U_{\mathtt F\cup\mathtt Q}\).
Its causal estimand is the vector
\((\Phi_{\mathtt F_1},\ldots,\Phi_{\mathtt F_d},
\Phi_{\mathtt Q_1},\ldots,\Phi_{\mathtt Q_d})\).

\paragraph{Report variants and granularity.}
For an event \(x\), let \(e^{(r)}(x;F^{(r)})\) be its exogenous state when abducted through regime \(r\)'s invertible equations, and let \(\delta_{e^{(r)}}\) denote the corresponding point-mass law.
The same four-object construction supports at least the three causal questions in Table~\ref{tab:causal-questions}.
We denote their estimands by \(\Phi_u\), \(\Phi_u^{\mathrm{shift}}(e^{(1)})\), and \(\Phi_u^{\mathrm{fix}}(e^{(0)})\), respectively.
\begin{table*}[t]
\centering
\small
\setlength{\tabcolsep}{3pt}
\renewcommand{\arraystretch}{1.05}
\begin{tabularx}{\textwidth}{@{}
>{\raggedright\arraybackslash}p{0.085\textwidth}
>{\raggedright\arraybackslash}p{0.14\textwidth}
>{\raggedright\arraybackslash}p{0.12\textwidth}
>{\raggedright\arraybackslash}p{0.14\textwidth}
>{\raggedright\arraybackslash}p{0.17\textwidth}
>{\raggedright\arraybackslash}p{0.07\textwidth}
>{\raggedright\arraybackslash}X@{}}
\toprule
Report & Causal question & Source & Target & Contrast & Players & Causal estimand \\
\midrule
Population shift
& Which regime changes moved the mean score?
& \(M(F^{(0)},Q^{(0)})\)
& \(M(F^{(1)},Q^{(1)})\)
& \(T(M^{(1)})-T(M^{(0)})\)
& \(\mathcal U_{\mathtt F\cup\mathtt Q}\)
& \(\Phi_u\): hybrid expected-score game \\
\midrule
Shifted event
& Which target-state components explain this event?
& \(M(F^{(0)},Q^{(0)})\)
& \(M(F^{(1)},\delta_{e^{(1)}})\)
& \(\tau(x)-T(M^{(0)})\)
& \(\mathcal U_{\mathtt F\cup\mathtt Q}\)
& \(\Phi_u^{\mathrm{shift}}(e^{(1)})\): abducted-event game \\
\midrule
Fixed-SCM event
& Which disturbances explain the event under \(M^{(0)}\)?
& \(M(F^{(0)},Q^{(0)})\)
& \(M(F^{(0)},\delta_{e^{(0)}})\)
& \(\tau(x)-T(M^{(0)})\)
& \(\mathcal U_{\mathtt Q}\)
& \(\Phi_u^{\mathrm{fix}}(e^{(0)})\): fixed-function event game \\
\bottomrule
\end{tabularx}
\caption{Three representative causal score-attribution questions in the source--target--player--score framework.}
\label{tab:causal-questions}
\end{table*}
The two event rows use different model-relative abductions: the fixed-SCM estimand holds \(F^{(0)}\) fixed, whereas the shifted-event estimand uses \(F^{(1)}\).
Player granularity is part of the estimand.
We use the flat function/law player set as primary; Supplement Secs.~A.1--A.2 give closest-work reductions and a hierarchy-preserving paired-node/Owen variant \citep{owen1977values}.

\paragraph{Causal replacement semantics.}
\label{subsec:state-replacement}
Activating \(\mathtt F_j\) replaces the autonomous parent-to-node equation; activating \(\mathtt Q_j\) replaces the direct exogenous law while holding that equation fixed.
The causal interpretation requires modularity, independent aligned disturbances, and common variable and residual coordinates across regimes.
It can fail under latent confounding or regime-specific sensor definitions.

\begin{proposition}[Efficiency and event--population coherence]
\label{prop:direct-population-coherence}
Let \(\Phi_u\) denote the population-shift estimand above and \(\Phi_u^{\mathrm{shift}}(e)\) its shifted-event counterpart, constructed with the same player set \(\mathcal U\) and replacement semantics.
Suppose the grand coalition of the population game reaches \(M^{(1)}\).
For integrable coalition values and \(e\sim Q^{(1)}\),
\begin{align}
\sum_{u\in\mathcal U}
\Phi_u&=T(M^{(1)})-T(M^{(0)}),
\label{eq:general-module-efficiency}
\\
\mathbb E_{e\sim Q^{(1)}}[\Phi_u^{\mathrm{shift}}(e)]
&=\Phi_u
\qquad\text{for every }u\in\mathcal U.
\label{eq:general-event-population-coherence}
\end{align}
\end{proposition}
The efficiency identity makes the population diagnosis complete for the selected score contrast: its component attributions account for the entire mean-score change.
Event--population coherence makes the two causal estimands directly comparable: averaging shifted-event attributions over incident events recovers each population attribution exactly, provided the endpoints, players, and replacement semantics agree.
The proof is in Supplement Sec.~C.1; the fixed-SCM event corollary is in Sec.~C.3.

\paragraph{Relation to causal attribution and shift discovery.}
\label{sec:related-work}
The source--target--player--score framework makes the distinctions from related methods explicit.
Feature and fixed-SCM event attribution explain inputs, disturbances, or interventions under one supplied model, rather than changes between regime SCMs \citep{lundberg2017unified,budhathoki2022causal,heskes2020causal,strobl2022identifying,nguyen2024noisy}.
do-Shapley attributes interventional outcome changes to variable-intervention players within a fixed SCM, not function/law replacements between regime SCMs \citep{jung2022measuring,witter2026exactly}.
Unit-level change attribution decomposes individual outcome differences,
whereas causal change attribution (CCA) and multiply-robust CCA (MRCCA)
replace node-wise generating distributions; neither provides function/law-resolved
population alarm-score attribution
\citep{budhathoki2022unit,budhathoki2021why,zhang2023model,quintasmartinez2024multiply}.
Mechanism-shift and difference-DAG methods detect shifted mechanisms or structural differences, or leverage shifts to recover causal structure, but do not allocate a supplied score contrast after causal propagation \citep{huang2020causal,perry2022causal,wang2018direct,malik2024identifying,chen2023iscan}.
Source-to-target regression transfers predictive models but does not define or compute this causal attribution estimand \citep{bastani2021predicting,takada2020transfer,li2022translasso}.
S$^4$-RCA instead meets the resolution, computation, and scarce-incident requirements jointly by separating function/law players, computing their full attribution exactly in polynomial time, and estimating it through reference-anchored transfer.
Supplement Table~1 (Sec.~A.1) gives the detailed estimand-level mapping and distinctions.

\paragraph{Why function/law resolution matters.}
Separating function and law effects matters for intervention and repair because
the two can favor different actions at the same node.  In a controlled oracle
construction, a target \(t\) has effects
\((\Phi_{\mathtt F_t},\Phi_{\mathtt Q_t})=(-\beta,+\alpha)\) and a distractor
has law effect \(\delta\), where
\(\alpha>\beta>\delta>\alpha-\beta>0\).  Pairing leaves the target below the
distractor, although restoring its law clears the alarm (Supplement Sec.~D.2).
FANS identifies function versus noise shifts in broad nonlinear SCMs
\citep{seo2026fans}, but does not attribute their propagated contributions to
a supplied score contrast; the two tasks are complementary.

\section{Exact Second-Order Causal Attribution}
\label{sec:causal-shapley}
\label{sec:mechanism-change-analysis}

\paragraph{Second-order specialization.}
\label{subsec:linear-scm}
To support immediate response, we specialize the Section~\ref{sec:prelim}
estimand to ordered linear SCMs and quadratic scores (polynomials of degree at
most two).  Their population
coalition values depend only on first and second exogenous moments.  The
estimand remains
\(\Phi_u(M^{(0)},M^{(1)},\mathcal U,T)\), with
\(M^{(r)}=M(B^{(r)},Q^{(r)})\), \(r\in\{0,1\}\), satisfying
\begin{align}
x=(B^{(r)})^\top x+e,\qquad e\sim Q^{(r)},
\label{eq:scm}
\end{align}
where \(B^{(r)}\) is strictly upper triangular in the known causal order and the exogenous coordinates are independent.
There is no separate structural intercept: \(\mathtt F_j\) switches the column \(B^{(r)}_{:,j}\), whereas \(\mathtt Q_j\) switches the law of \(e_j\), including its location.
Let \(\mu_j^{(r)}=\mathbb E_{Q^{(r)}}[e_j]\), \(\omega_j^{(r)}=\mathrm{Var}_{Q^{(r)}}(e_j)\), and
\begin{align}
A^{(r)}=(I-B^{(r)})^{-1},\qquad x=(A^{(r)})^\top e.
\label{eq:mixing}
\end{align}
The supplied anomaly score is
\label{sec:anomaly-scores}
\begin{align}
\tau(x)=x^\top W x+z^\top x.
\label{eq:tau}
\end{align}
We omit additive constants, which cancel from all coalition marginals and
endpoint contrasts, and take \(W=W^\top\); otherwise replace it by
\((W+W^\top)/2\).  In reference-fitted coordinates, this includes
PCA \(T^2\), ICA \(I^2\), and SPE; quadratic scores formed from Mahalanobis
or graphical-lasso precision matrices; raw or squared regression residuals; squared \(z\)-scores; and
individual coordinates \(x_j\).
Thus this covers standard multivariate monitors while requiring only incident
first and second moments, rather than an incident density estimate.
Here \(\Diag(v)\) forms the diagonal matrix with diagonal \(v\), and \(\diag(M)\) extracts the diagonal vector of \(M\).
For any endpoint or hybrid state, suppress state superscripts and let
\(\Omega=\Diag(\omega)\), \(V=AWA^\top\), and \(\ell=Az\).
Since \(x=A^\top e\),
\begin{align}
\tau(A^\top e)=e^\top V e+\ell^\top e.
\label{eq:residual-quadratic-score}
\end{align}
Taking expectations gives the closed-form target functional:
\begin{align}
T(M(B,Q))
&=\mathrm{tr}(V\Omega)+\mu^\top V \mu+\ell^\top \mu.
\label{eq:mean-intercept-population-game}
\end{align}
We also write this same value as \(T(B,\mu,\omega)\) when the linear endpoint parameters are explicit.

\paragraph{Hybrid values.}
\label{subsec:distributional-attribution}
\label{subsec:linear-population-shapley}
We use the primary function/law-resolved granularity
\(\mathcal U=\mathcal U_{\mathtt F\cup\mathtt Q}\); the construction below
also applies to other node-local granularities.
For \(U\subseteq\mathcal U\), the hybrid quantities determining
\(T(M^U)=T(B^U,\mu^U,\omega^U)\) are
\begin{align}
B^U_{:,j}
&=B^{(1)}_{:,j}\mathbf 1\{\mathtt F_j\in\bigcup_{w\in U}w\}
 +B^{(0)}_{:,j}\mathbf 1\{\mathtt F_j\notin\bigcup_{w\in U}w\},\notag\\
\mu^U_j
&=\mu_j^{(1)}\mathbf 1\{\mathtt Q_j\in\bigcup_{w\in U}w\}
 +\mu_j^{(0)}\mathbf 1\{\mathtt Q_j\notin\bigcup_{w\in U}w\},\notag\\
\omega^U_j
&=\omega_j^{(1)}\mathbf 1\{\mathtt Q_j\in\bigcup_{w\in U}w\}
 +\omega_j^{(0)}\mathbf 1\{\mathtt Q_j\notin\bigcup_{w\in U}w\}.
\label{eq:linear-population-hybrid}
\end{align}
The game is simulation-free, but direct Shapley evaluation still requires all
\(K!\) player orderings or all \(2^K\) coalitions, where \(K=|\mathcal U|\).
Let \(R_p\subseteq\mathcal U\) include each module \(u\in\mathcal U\)
independently with probability \(p_u\).
The multilinear extension of \(U\mapsto T(M^U)\) is \citep{owen1972multilinear}
\begin{align}
\widetilde T(p)
&:=\mathbb E_{R_p}\!\left[T(M^{R_p})\right]
=\sum_{U\subseteq\mathcal U}T(M^U)\Pr(R_p=U),
\label{eq:main-multilinear-extension}
\end{align}
for \(p=(p_u)_{u\in\mathcal U}\in[0,1]^K\).
Each Shapley value can be represented by integrating a partial derivative of
this multilinear extension \citep{shapley1953value,owen1972multilinear}, but
the extension and its derivatives must still be evaluated without constructing
the exponential coalition table.

\paragraph{Moment recursion.}
Draw \(R_p\) and then \(x\) from the selected hybrid SCM, and define
\(\bar x(p)=\mathbb E[x]\) and \(S(p)=\mathbb E[xx^\top]\).
These random-switch moments evaluate \(\widetilde T\), with local recursion
inputs computed once from the endpoints.
Write \(b_j^{(r)}=B^{(r)}_{<j,j}\in\mathbb R^{j-1}\),
\(Z_{\mathtt F,j}=\mathbf 1\{\mathtt F_j\in\bigcup_{w\in R_p}w\}\), and
\(Z_{\mathtt Q,j}=\mathbf 1\{\mathtt Q_j\in\bigcup_{w\in R_p}w\}\).
Expectations below are over these switches:
\begin{align}
\bar b_j(p)&=\mathbb E[b_j^{Z_{\mathtt F,j}}],\quad
\bar\mu_j(p)=\mathbb E[\mu_j^{Z_{\mathtt Q,j}}],\notag\\
\Gamma_j(p)&=\mathbb E[b_j^{Z_{\mathtt F,j}}(b_j^{Z_{\mathtt F,j}})^\top],\quad
c_j(p)=\mathbb E[b_j^{Z_{\mathtt F,j}}\mu_j^{Z_{\mathtt Q,j}}],\notag\\
q_j(p)&=\mathbb E[(\mu_j^{Z_{\mathtt Q,j}})^2+\omega_j^{Z_{\mathtt Q,j}}].
\label{eq:random-switch-local-moments}
\end{align}
Because the SCM is ordered, the required moments can be propagated node by
node.

\begin{theorem}[Exact second-order moment recursion]
\label{thm:population-moment-recursion}
Assume the ordered source and target states have independent exogenous
coordinates with finite second moments, \(\mathcal U\) is node-local, and no
atomic component appears in two modules.
The coordinates of \(\bar x(p)\) and \(S(p)\) are computed in causal order by
the following recursion:
\begin{align}
\bar x_j(p)&=\bar b_j(p)^\top \bar x_{<j}(p)+\bar\mu_j(p),\notag\\
S_{<j,j}(p)&=S_{<j,<j}(p)\bar b_j(p)
 +\bar x_{<j}(p)\bar\mu_j(p),\notag\\
S_{jj}(p)&=\mathrm{tr}\{\Gamma_j(p)S_{<j,<j}(p)\}\notag\\
&\quad +2c_j(p)^\top\bar x_{<j}(p)+q_j(p).
\label{eq:main-moment-recursion}
\end{align}
Then, for every \(u\in\mathcal U\), the Shapley value is
\begin{align}
\Phi_u(M^{(0)},M^{(1)},\mathcal U,T)
&=\int_0^1
\left.\frac{\partial\widetilde T(p)}{\partial p_u}
\right|_{p=t\mathbf 1}\,dt,
\label{eq:main-shapley-multilinear-integral}
\end{align}
where
\begin{align}
\widetilde T(p)
&=\mathrm{tr}\{W S(p)\}+z^\top \bar x(p).
\label{eq:main-random-switch-target}
\end{align}
These equations compute every Shapley value without enumerating coalitions.
\end{theorem}

The key is second-order closure: random function/law replacements enter
through the local moments in~\eqref{eq:random-switch-local-moments}, and the
causal recursion propagates exactly the moments required by the score rather
than approximating the coalition game.

\begin{corollary}[Exact all-component computation]
\label{cor:exact-all-component}
For the primary function/law-resolved game, one forward evaluation of the recursion
in~\eqref{eq:main-moment-recursion} followed by one shared backward-sensitivity sweep returns
\(\nabla_p\widetilde T(p)\) for all \(K=2d\) players in \(O(d^3)\) dense time and \(O(d^2)\) memory.
Combining these gradients with \(d\)-point Gauss--Legendre quadrature computes the entire Shapley vector exactly up to numerical roundoff in
\(O(Kd^3)=O(d^4)\) time and \(O(d^2)\) memory.
\end{corollary}

At each of the \(d\) quadrature points, a shared forward--reverse sweep returns
all \(2d\) derivatives. The integrand has degree at most \(2d-1\), so
\(d\)-point Gauss--Legendre quadrature is exact \citep{golub1969calculation}.
Supplement Sec.~C.2 and Algorithm~1 give the reverse equations, proof, and
full procedure. The resulting speedup preserves the resolved estimand:
tractability does not require pairing function and law players.

\paragraph{Event corollary.}
Fixing \(F\) and replacing the target law by abducted point masses reduces the
game to the expected-score counterpart of \citet{budhathoki2022causal}: it
shares factual-noise replacement but uses \(\mathbb E_M[\tau(X)]\), rather than
an information-theoretic tail score.  Every \(F\) player is then null, so the
same recursion yields fixed-SCM event attribution without a separate algorithm
(Supplement Secs.~C.3 and A.1).

\section{Sample-Efficient Plug-in Estimation}
\label{sec:plugin-estimation}

Given the supplied score, causal order or parent sets, and player set, we
estimate the Section~\ref{sec:mechanism-change-analysis} target by plugging
sample estimates of \(B^{(r)},\mu^{(r)},\omega^{(r)}\) into the exact program.
For quadratic \(\tau\), only residual means and variances are needed beyond the
mechanism. We first bound reference-anchored transfer and then propagate
endpoint errors to the Shapley vector, using held-out or out-of-fold residuals
for concentration.

\paragraph{Identification for the linear plug-in.}
Under~\eqref{eq:scm}, a correct common order (or parent sets)
and nonsingular parent covariances identify \(B^{(r)}_{:,j}\) as population
regression slopes; exogenous independence identifies the residual laws
\(Q_j^{(r)}\) and justifies product-law hybrids. Physical function/law
interpretation additionally requires autonomous equations and stable variable
semantics. Thus aligned endpoint states and semantics are inputs to, rather
than outputs of, S$^4$-RCA.

\paragraph{Incident adaptation from a reference SCM.}
\label{subsec:mca-estimators}
To estimate \(B^{(1)}\) when \(n_1\) is small, we fit the ordered SCM to the
large pre-incident sample and use incident data only to estimate its deviation.
All fits and residual moments use a common reference-fitted coordinate system;
centering and intercept bookkeeping assign residual location to \(Q\), as
detailed in Supplement Algorithm~2.
Let \(\mathbf X^{(r)}\) denote the resulting centered or
intercept-residualized regression data.
Let \(\widetilde b_k\) be the pre-incident nodewise Lasso estimate for node \(k\).
Delta-Lasso fits
\begin{align}
\widehat\delta_k
&\in\argmin_\delta\left\{\frac{1}{2n_1}\|\mathbf y_k-\mathbf X^{(1)}_{:,<k}\delta\|_2^2+\lambda_1\|\delta\|_1\right\},
\notag\\
\mathbf y_k&=\mathbf X^{(1)}_{:,k}-\mathbf X^{(1)}_{:,<k}\widetilde b_k,
\qquad \widehat b_k^\delta=\widetilde b_k+\widehat\delta_k.
\label{eq:mca-delta-reparam}
\end{align}
For downstream plug-ins, set \(\widehat B^{(0)}:=\widetilde B\) and \(\widehat B^{(1)}:=\widehat B^\delta\).
Write \(\mathbf Z_k^{(r)}=\mathbf X_{:,<k}^{(r)}\) and
\(\mathcal C(S;3)=\{\theta:\|\theta_{S^c}\|_1\le3\|\theta_S\|_1\}\).
Here \(S^c\) is the complement within the nodewise regression coordinates;
\(\|\cdot\|_{\mathrm{op}}\), \(\|\cdot\|_{\mathrm F}\), and
\(\|\cdot\|_{1,1}\) denote the operator, Frobenius, and elementwise
\(\ell_1\) matrix norms.  We use \(O(\cdot),o(\cdot)\) for deterministic
orders and \(O_{\mathrm P}(\cdot),o_{\mathrm P}(\cdot)\) for stochastic
orders. The column-normalization and conditional sub-Gaussian conditions with
scale \(\sigma_r\) in Supplement Sec.~C.4 yield one joint score event over all
nodewise regressions; that section also gives random-design sufficient
conditions, while Sec.~C.5 gives the proofs.

\begin{assumption}[Reference mechanism fit \textup{(M0)}]
\label{ass:reference-fit}
Reference rows are independent.
With \(S_k^{(0)}=\operatorname{supp}(b_k^{(0)})\), assume \(\sum_k|S_k^{(0)}|=s_0\) and a common \(\kappa_0>0\) such that
\(n_0^{-1}\|\mathbf Z_k^{(0)}\theta\|_2^2\ge\kappa_0\|\theta\|_2^2\)
for every \(k\in[d]\) and \(\theta\in\mathcal C(S_k^{(0)};3)\).
\end{assumption}

\begin{assumption}[Incident transfer \textup{(M1)}]
\label{ass:incident-transfer}
Incident rows are mutually independent and independent of the reference sample.
For common \(\kappa_1>0\) and \(\kappa_2\ge0\),
\(n_1^{-1}\|\mathbf Z_k^{(1)}\theta\|_2^2
\ge\kappa_1\|\theta\|_2^2-\kappa_2(\log d/n_1)\|\theta\|_1^2\)
for every \(k\in[d]\) and \(\theta\in\mathbb R^{k-1}\).
\end{assumption}

For the residual-moment plug-in below, we additionally use the following
scaling condition.
\begin{assumption}[Residual-moment scaling \textup{(R)}]
\label{ass:residual-moments}
Uniformly over \(r\in\{0,1\}\), assume
\(\sigma_r=O(1)\) and \(\|A^{(r)}\|_{\mathrm{op}}=O(1)\).
\end{assumption}

\begin{theorem}[Radius-based SCM Delta-Lasso transfer]
\label{thm:mca-delta-core}
Under the column-normalization and sub-Gaussian conditions above and
Assumptions~\ref{ass:reference-fit} and~\ref{ass:incident-transfer}, let
\(\Delta=B^{(1)}-B^{(0)}\), choose
\(\lambda_r\asymp\sigma_r\sqrt{\log d/n_r}\), and set
\begin{align}
R_\Delta
&:=\|\Delta\|_{1,1}+\frac{12\lambda_0s_0}{\kappa_0}.
\label{eq:main-delta-radius}
\end{align}
For any fixed \(c>0\), suppose
\(\kappa_2R_\Delta\sqrt{\log d/n_1}\le c\sigma_1\).
Then, with high probability,
\begin{align}
\|\widehat B^\delta-B^{(1)}\|_{\mathrm F}^2
&\le C R_\Delta
\frac{\sigma_1}{\kappa_1}\sqrt{\frac{\log d}{n_1}},
\label{eq:mca-delta-core-bound}
\end{align}
where \(C\) does not depend on \(n_0,n_1\), or \(d\).
\end{theorem}

\paragraph{When transfer helps.}
Let \(s_1=\|B^{(1)}\|_0\) count the incident nonzero slopes.  Under an
incident nodewise RE constant \(\kappa>0\), standard incident-only Lasso has Frobenius rate
\(O_{\mathrm P}((\sigma_1/\kappa)\sqrt{s_1\log d/n_1})\).
Squaring this rate and comparing guarantees, transfer is sharper when
\(R_\Delta\) is
\(o\!\left(
(\sigma_1\kappa_1/\kappa^2)s_1
\sqrt{\log d/n_1}
\right)\).
With comparable noise and curvature constants, \(s_0=O(s_1)\), and
\(n_0/n_1\to\infty\), the reference-estimation term is negligible and the
condition reduces to
\(\|\Delta\|_{1,1}=o(s_1\sqrt{\log d/n_1})\): the favorable regime combines
\(n_0\gg n_1\) with a small total shift.  The shift may still be dense:
the condition controls \(\|\Delta\|_{1,1}\), not its support size.

\paragraph{Held-out residual moments.}
\label{subsec:plugin-stability}
For a regime-\(r\) residual-evaluation fold \(\ell\), let
\(\mathcal I_{r,\ell}\) be its held-out observations,
\(m_{r,\ell}=|\mathcal I_{r,\ell}|\), and \(\widehat B_{-\ell}^{(r)}\) the fit
excluding that fold; we suppress \(\ell\) below.
\begin{align}
\widehat{\mathrm e}_i^{(r)}
&=(I-\widehat B_{-\ell}^{(r)\top})x_i^{(r)},\qquad i\in\mathcal I_r,
\notag\\
(\widehat\mu_j^{(r)},\widehat\omega_j^{(r)})
&=m_r^{-1}\sum_{i\in\mathcal I_r}
(\widehat{\mathrm e}_{ij}^{(r)},(\widehat{\mathrm e}_{ij}^{(r)}-\widehat\mu_j^{(r)})^2).
\label{eq:omega-hat}
\end{align}
Excluding \(\mathcal I_{r,\ell}\) makes its residuals independent of
\(\widehat B_{-\ell}^{(r)}\). Each fold forms matched mechanism and
residual-moment endpoints and a Shapley vector in common coordinates. The
report averages these vectors rather than pooling residuals across games with
different mechanism fits.

\paragraph{Plug-in attribution and ranking.}
For a node-local game, replace the endpoints in
\eqref{eq:linear-population-hybrid} to form
\(\widehat B^U,\widehat\mu^U,\widehat\omega^U\), with plug-in value
\begin{align}
\widehat T(U)
&=T(\widehat B^U,\widehat\mu^U,\widehat\omega^U).
\label{eq:population-plugin-game}
\end{align}
Let \(\widehat\Phi=(\widehat\Phi_u)_{u\in\mathcal U}\) be the Shapley vector
obtained by applying~\eqref{eq:causal-shapley} to
\(U\mapsto \widehat T(U)\).
Put
\(\Theta=(\theta^{(0)},\theta^{(1)})\), where
\(\theta^{(r)}=(B^{(r)},\mu^{(r)},\omega^{(r)})\), and let
\(\widehat\Theta\) be its plug-in counterpart.  Regard
\(\Phi_u(\Theta)\) as an endpoint functional, and let
\(\Theta_s=\Theta+s(\widehat\Theta-\Theta)\), \(s\in[0,1]\).
Let \(\mathcal B\) be the prespecified admissible slope subspace (the supplied
order or parent support) containing the fitted and population slopes, with
Frobenius projection \(\Pi_{\mathcal B}\).
Write
\(\eta_B^{(r)}=\|\widehat B^{(r)}-B^{(r)}\|_{\mathrm F}\),
\(\eta_\mu^{(r)}=\|\widehat\mu^{(r)}-\mu^{(r)}\|_\infty\), and
\(\eta_\omega^{(r)}=\|\widehat\omega^{(r)}-\omega^{(r)}\|_\infty\).
The following path sensitivities measure how endpoint errors can perturb
player \(u\)'s attribution along the segment from the population to the
fitted endpoints:
\begin{align}
L_{u,B}^{(r)}
&:=\sup_{s\in[0,1]}\|\Pi_{\mathcal B}
\nabla_{B^{(r)}}\Phi_u(\Theta_s)\|_{\mathrm F},
\notag\\
L_{u,\mu}^{(r)}
&:=\sup_{s\in[0,1]}\|\nabla_{\mu^{(r)}}
\Phi_u(\Theta_s)\|_1,\notag\\
L_{u,\omega}^{(r)}
&:=\sup_{s\in[0,1]}\|\nabla_{\omega^{(r)}}
\Phi_u(\Theta_s)\|_1.
\label{eq:player-specific-sensitivity-envelopes}
\end{align}

\paragraph{Uniform error rate.}
For deterministic \(\rho_r\) satisfying
\(\eta_B^{(r)}=O_{\mathrm P}(\rho_r)\), put
\(\nu_r=\sqrt{\log(2d)/m_r}\),
\(M_r=\|\mathbb E x^{(r)}\|_2\),
\(q_{\mu,r}=\nu_r+(M_r+\nu_r)\rho_r\),
\(q_{\omega,r}=\nu_r+\rho_r^2\), and
\(a_*=\max_{U\subseteq\mathcal U}\|(I-B^U)^{-1}\|_{\mathrm{op}}\).
With \(W_*=\|W\|_{\mathrm{op}}\) and \(z_*=\|z\|_2\), define the
deterministic envelopes below, which may depend on \(d\):
\begin{align*}
m_*&=1+\sum_{r=0}^1\{\|\mu^{(r)}\|_2+\sqrt d\,q_{\mu,r}\},\\
\omega_*&=1+\max_r\{\|\omega^{(r)}\|_\infty+q_{\omega,r}\},\\
H_B&=16a_*^3(\sqrt d\,\omega_*+m_*^2)W_*+4a_*^2m_*z_*,\\
H_\mu&=\sqrt d\{8a_*^2W_*m_*+2a_*z_*\},\quad
H_\omega=4da_*^2W_*.
\end{align*}

\begin{theorem}[Plug-in attribution error]
\label{thm:population-plugin-rate}
For any fitted endpoints with slopes in \(\mathcal B\), every player obeys
the deterministic sensitivity bound
\begin{align}
|\widehat\Phi_u-\Phi_u|
&\le\sum_{r=0}^1\{L_{u,B}^{(r)}\eta_B^{(r)}
+L_{u,\mu}^{(r)}\eta_\mu^{(r)}
+L_{u,\omega}^{(r)}\eta_\omega^{(r)}\}.
\label{eq:player-specific-deterministic-bound}
\end{align}
For the out-of-sample construction above, under the held-out-sample
independence and sub-Gaussian conditions above and
Assumption~\ref{ass:residual-moments}, if
\(\max_r\nu_r=o(1)\) and \(a_*(\rho_0+\rho_1)=o(1)\), then, simultaneously
over the prespecified player universe,
\(\|\widehat\Phi-\Phi\|_\infty=O_{\mathrm P}(R_d)\), where
\begin{align}
R_d&:=\sum_{r=0}^1\{H_B\rho_r+H_\mu q_{\mu,r}+H_\omega q_{\omega,r}\}.
\label{eq:player-specific-plugin-rate}
\end{align}
For a fixed number of folds, averaging their foldwise bounds preserves the
corresponding order.
\end{theorem}
The theorem converts endpoint rates into attribution accuracy. For the
estimators above, take
\(\rho_0=(\sigma_0/\kappa_0)\sqrt{s_0\log d/n_0}\), while
Theorem~\ref{thm:mca-delta-core} gives
\(\rho_1=[R_\Delta
(\sigma_1/\kappa_1)\sqrt{\log d/n_1}]^{1/2}\).
Reference accuracy and a small total shift reduce these rates, giving uniform
consistency whenever \(R_d=o(1)\); endpoint errors or hybrid amplification
through \(a_*\) can negate the benefit. Even when \(a_*,W_*=O(1)\), the
\(H_\omega q_{\omega,r}\) term in the dense-score global
envelope is \(O_{\mathrm P}(d\sqrt{\log d/m_r}+d\rho_r^2)\) and can require
\(m_r\gg d^2\log d\), whereas the player-specific
bound~\eqref{eq:player-specific-deterministic-bound} retains cancellations.

\begin{corollary}[Top-\(k\) recovery under a separated ranking]
\label{cor:ranking-recovery}
Under the conditions of Theorem~\ref{thm:population-plugin-rate},
let \(S_k\) be the population top-\(k\) set and
\(g_k:=\min_{u\in S_k}|\Phi_u|-\max_{v\notin S_k}|\Phi_v|\).
If \(g_k>0\) and \(R_d=o(g_k)\),
the estimated and population top-\(k\) absolute-attribution sets agree with probability tending to one.
For any \(u\in\mathcal U\), \(R_d=o(|\Phi_u|)\) also gives sign recovery.
\end{corollary}
Thus the leading function/law attributions are ranked correctly once their
population gap dominates the uniform estimation error.

\begin{figure}[t]
  \centering
  \includegraphics[width=\linewidth]{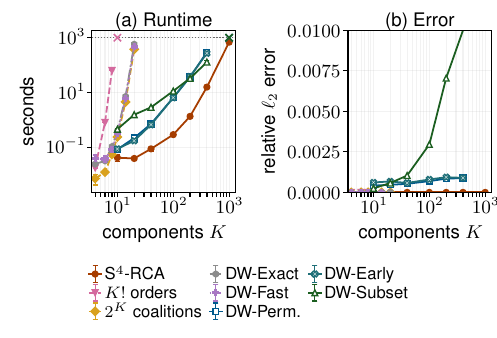}
  \caption{Same-oracle Shapley aggregation on the full \(K=2d\) function/law game (mean \(\pm\) SE over three non-timeout problems; crosses: first 1000-second timeout). Only exact S$^4$-RCA reaches \(K=1000\). Panel (b) reports relative \(\ell_2\) error; DW-Fast's diagnosed \(K=20\) failure is omitted (DoWhy 0.14; Supplement Sec.~D.3).}
  \label{fig:main-scalability}
\end{figure}

\begin{figure}[t]
  \centering
  \includegraphics[width=\linewidth]{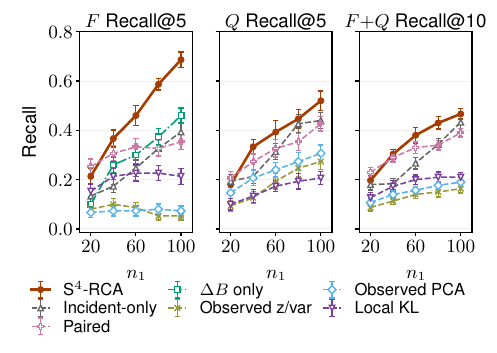}
  \caption{Scarce-incident recovery over 30 ordered \(d=100\) SCM pairs with
  balanced function/law oracle mass (mean \(\pm\) SE). Panels report \(F\) and
  \(Q\) Recall@5 and \(F+Q\) Recall@10; node rankings are lifted for \(F+Q\).
  Beyond the smallest regime, S$^4$-RCA has the highest mean, most clearly for
  \(F\) (broader benchmark: Supplement Sec.~D.4).}
  \label{fig:component-benchmark}
\end{figure}

\section{Experiments}
\label{sec:experiments}

We test exact computation, scarce-incident function/law recovery, and
model-relative measured-intervention diagnosis. Supplement Secs.~D.1--D.8 give
protocols, uncertainty summaries, stress tests, and reproducibility details.

\paragraph{Exact computation.}
\label{sec:exp_scalability}
All methods query the same oracle for true-endpoint coalition values. We use
DoWhy-GCM for the DW comparators \citep{blobaum2024dowhygcm};
Figure~\ref{fig:main-scalability} compares runtime and relative \(\ell_2\) error.
Across three problems per non-timeout cell,
S$^4$-RCA is over \(100\times\) faster than \(K!\) orders, \(2^K\) coalitions,
DW-Exact, and DW-Fast at their largest common sizes; DW-Fast also loses
agreement at \(K=20\). Only S$^4$-RCA reaches \(K=1000\) within \(10^3\)
seconds, averaging about 11 minutes (Supplement Sec.~D.3).

\paragraph{Scarce-incident attribution benchmark.}
\label{sec:exp_component_benchmark}
We use 30 ordered Gaussian \(d=100\) SCM pairs with five score-relevant shifts
per family, matched score-null decoys, and balanced oracle mass.
S$^4$-RCA averages signed attributions over five outer folds; ten-fold penalty
tuning uses only each mechanism-training fold.
Figure~\ref{fig:component-benchmark} reports \(F\) and \(Q\) Recall@5 and
\(F+Q\) Recall@10. Incident-only Lasso, paired attribution, and
\(\lvert\Delta B\rvert\) ablate transfer, resolution, and score-aware
propagation; observed baselines provide diagnostic anchors.

The advantage is clearest for function recovery: at the largest incident size,
recall is \(0.687\), versus \(0.460\) for \(\lvert\Delta B\rvert\) and \(0.393\)
for incident-only Shapley. Paired attribution is higher only in the smallest
regime; thereafter S$^4$-RCA has the highest mean on all three metrics, with
higher \(F+Q\) Recall than incident-only throughout and higher \(Q\) Recall
beyond the smallest regime. Thus transfer and score propagation chiefly
improve function recovery. This matches the intended transfer setting: a
reliable reference anchors the dense baseline. A lifted node report can score similarly on
\(F+Q\) without identifying function versus law, so the resolved report remains
useful even when untyped localization is comparable. Supplement Sec.~D.4 gives
the protocol and a broader benchmark including fixed-\(F\) event attribution,
CCA/MRCCA, DCI, iSCAN, and feature-shift SHAP. The evidence is for leading-component
recovery, not uniform magnitude recovery or calibrated ranks.

\paragraph{Measured interventions.}
Across 16 22-channel Causal Chamber Light-Tunnel interventions
\citep{gamella2025chamber}, we test whether a measured target-law component
separates from propagated fitted-response components of a reference-fitted PCA
\(T^2+\mathrm{SPE}\) alarm. Five repeated two-fold fits define a 44-player game,
and nodes are ranked by the sum of their absolute fold-mean function and law
attributions. Fixed-\(F\) and observed PCA ablate function adaptation and
causal propagation.

\begin{figure}[!t]
  \centering
  \includegraphics[width=\linewidth]{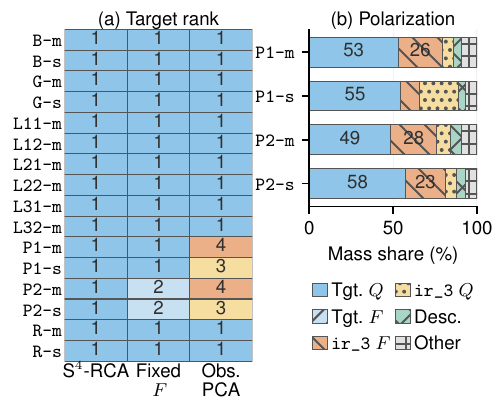}
  \caption{Causal Chamber all-channel score. (a) Target rank (1 best; worst of five splits; \texttt{m}/\texttt{s}: severity). (b) Target-\(Q\) and downstream-\texttt{ir\_3} mass; \(F\) denotes the fitted linear mechanism.}
  \label{fig:causal-chamber-case}
\end{figure}

S$^4$-RCA ranks all 16 targets first, versus 14 for Fixed-\(F\) (all top-three)
and 12 for observed PCA. For polarization, target \(Q\) carries
\(49\)--\(58\%\) of absolute mass, while \(F\) carries \(14\)--\(32\%\),
\(91\)--\(100\%\) downstream; downstream function/law effects oppose at mid
severity and reinforce at strong severity (Figure~\ref{fig:causal-chamber-case}).
Within the fitted ordered linear SCM, this allocation separates the target-law
component from downstream relinearization. Supplement Sec.~D.6 gives the full
protocol and split sensitivity; Sec.~D.7 extends the stress test to two other
datasets.

\section{Conclusion and Limitations}
\label{sec:limitations}
S$^4$-RCA combines resolved attribution, exact computation, and scarce-incident
estimation. Its guarantees assume aligned ordered linear SCMs with independent
disturbances and quadratic scores; transfer additionally requires an accurate
reference and a small shift. Latent confounding, cycles, and incompatible
semantics remain outside scope. Small-scale stress tests suggest rank robustness
to order estimation and controlled nonlinear misspecification but do not extend
the theory (Supplement Sec.~D.5). Measured results are model-relative;
physical-fault diagnosis requires benchmarks with verified mechanisms.

\clearpage
\begin{center}
\Large\bfseries
Supplementary Material for\\
What Moved the Alarm Score?\\
Causal Attribution with Exact Computation\\
and Scarce-Incident Estimation
\end{center}
\appendix
\numberwithin{equation}{section}
\renewcommand{\theequation}{\thesection-\arabic{equation}}
\setcounter{figure}{0}
\setcounter{table}{0}
\setcounter{algorithm}{0}
\setcounter{axiom}{0}
\renewcommand{\theHfigure}{supplement.\arabic{figure}}
\renewcommand{\theHtable}{supplement.\arabic{table}}
\providecommand{\theHalgorithm}{\arabic{algorithm}}
\renewcommand{\theHalgorithm}{supplement.\arabic{algorithm}}
\providecommand{\theHaxiom}{\arabic{axiom}}
\renewcommand{\theHaxiom}{supplement.\arabic{axiom}}

\noindent\textbf{Appendix organization.}
Appendix~\ref{app:general-positioning} gives the framework scope, closest-work reductions, and hierarchy-preserving Owen extension.
Appendix~\ref{app:algorithms} gives pseudocode and implementation details.
Appendix~\ref{app:proofs} proves the general identities, the linear/quadratic computation results, and the plug-in estimation guarantees used in Sections~\ref{sec:prelim}--\ref{sec:plugin-estimation}.
Appendix~\ref{sec:app_experiments} records the experimental protocols and boundary diagnostics needed to reproduce and interpret the main claims.

\section{Extensions and Closest-Work Reductions}
\label{app:general-positioning}

This appendix develops two aspects compressed in the main paper.  First, it
states the scope of the source--target--player--score construction and maps
closely related methods into that common notation.  It then gives an Owen-value
extension that preserves paired-node hierarchy while retaining separate
function/law allocations.

\subsection{Framework scope and closest-work reductions}
\label{app:close-work-positioning}

The Shapley definition in Eq.~\eqref{eq:causal-shapley} applies to any
finite scalar functional \(T\), whereas event--population coherence, moment
closure, and the estimation theory require the expected-score form
\(T(M)=\mathbb E_M[\tau(X)]\).

A function/law-resolved report is conditional on aligned variables and
exogenous coordinates across its endpoints; unrestricted observational laws
do not identify a unique physical function/noise decomposition.  The endpoints
may be supplied or estimated under additional assumptions.  In the ordered
linear specialization, a correct common order (or parent sets), nonsingular
parent covariances, and independent disturbances identify the slopes and
residual laws and justify product-law hybrids.  There is no separate structural
intercept: \(\mathtt F_j\) switches the slope, while \(\mathtt Q_j\) includes
residual location and dispersion.

Any attribution report is determined by four choices: source
\(M^{(0)}\), target \(M^{(1)}\), prespecified switches \(\mathcal U\), and
the scalar \(T\) being explained.  To prevent method-specific notation from
obscuring these choices, Table~\ref{tab:close-work-map} translates every
framework into this paper's source--target--player--score notation.
For attribution methods, ``source'' and ``target'' mean the empty- and
grand-coalition endpoints \(M^\emptyset\) and \(M^{\mathcal U}\), so the
allocated contrast is \(T(M^{\mathcal U})-T(M^\emptyset)\); they need not be
the normal and incident states in chronological order.

\begin{table*}[!t]
\centering
\small
\setlength{\tabcolsep}{2.5pt}
\renewcommand{\arraystretch}{1.0}
\begin{tabularx}{\textwidth}{@{}
>{\raggedright\arraybackslash}p{0.135\textwidth}
>{\raggedright\arraybackslash}p{0.13\textwidth}
>{\raggedright\arraybackslash}p{0.13\textwidth}
>{\raggedright\arraybackslash}p{0.145\textwidth}
>{\raggedright\arraybackslash}p{0.115\textwidth}
>{\raggedright\arraybackslash}X@{}}
\toprule
Framework
& \(M^{(0)}\): source
& \(M^{(1)}\): target
& Modules \(\mathcal U\)
& Scalar \(T\)
& Report in words \\
\midrule
\textbf{S$^4$-RCA (ours)}
&
\(M(F^{(0)},Q^{(0)})\)
&
\(M(F^{(1)},Q^{(1)})\)
&
\(\mathcal U_{\mathtt F\cup\mathtt Q}\):
separate \(\mathtt F_j,\mathtt Q_j\)
&
\(T(M)=\mathbb E_M[\tau(X)]\)
&
Allocates a population alarm-score change to structural functions and
disturbance laws separately.
\\
\cmidrule{1-6}
Population CCA/MRCCA
\citep{budhathoki2021why,quintasmartinez2024multiply}
&
\(M(F^{(0)},Q^{(0)})\)
&
\(M(F^{(1)},Q^{(1)})\)
&
\(\mathcal U_{(\mathtt F,\mathtt Q)}\):
joint node switches
&
\(T(M)=\psi(P_Y^M)\)
&
Switches each node conditional jointly.  MRCCA estimates
\(T(M)=\mathbb E_M[h(Y)]\) multiply robustly; with \(h(Y)=\tau(X)\), the
game is paired-node S$^4$-RCA.
\\
\cmidrule{1-6}
Unit-level change attribution
\citep{budhathoki2022unit}
&
\(M(F_Y^{(0)},\delta_{x^{(0)}})\)
&
\(M(F_Y^{(1)},\delta_{x^{(1)}})\)
&
\(\{\mathtt F_Y,\mathtt X_1,\ldots,\mathtt X_d\}\):
function/value switches
&
\(T(M)=\mathbb E_M[Y]=F_Y(x)\)
&
Explains one unit's output change by switching the prediction function and
observed input coordinates (fine game), or all inputs jointly (coarse game).
\\
\cmidrule{1-6}
Fixed-SCM outlier RCA
\citep{budhathoki2022causal}
&
\(M(F^{(0)},Q^{(0)})\)
&
\(M(F^{(0)},\delta_{e^{(0)}(x)})\)
&
\(\mathcal U_{\mathtt Q}\):
factual-noise fixes
&
\(T_x(M)=\log P_M(A_x)\)
&
Fixes reference noises to their factual values and allocates the
information-theoretic score \(S^\tau_{M^{(0)}}(x)\).
\\
\cmidrule{1-6}
FANS
\citep{seo2026fans}
&
\(M(F^{(0)},Q^{(0)})\)
&
\(M(F^{(1)},Q^{(1)})\)
&
---; returns \(\mathtt F_j/\mathtt Q_j\) labels
&
---
&
From samples in two environments, detects changed nodes and labels function
or noise shifts; it does not allocate a score contrast.
\\
\cmidrule{1-6}
Exact do-Shapley
\citep{witter2026exactly}
&
\(M^{(0)}=M\)
&
\(M^{(1)}=M_{\operatorname{do}(X=x)}\)
&
\(\{\mathtt X_j:j\in[d]\}\):
variable interventions
&
\(T(M)=\mathbb E_M[Y]\)
&
Allocates an interventional outcome to variable interventions, not
function/law shifts.
\\
\bottomrule
\end{tabularx}
\caption{Closest frameworks translated into this paper's notation.
CCA/MRCCA neither identifies nor separately switches the function/law
representation of an induced node conditional.  For outlier RCA,
\(A_x=\{\tau(X)\geq\tau(x)\}\) and
\(S^\tau_{M^{(0)}}(x)=-\log P_{M^{(0)}}(A_x)\).
Unit-level attribution observes paired input vectors rather than estimating
input laws.  Its deterministic one-output SCM embedding uses
\(\delta_{x^{(r)}}=\bigotimes_{j=1}^d\delta_{x_j^{(r)}}\), so
\(\mathtt X_j\) switches an observed value, not a population disturbance law.
FANS has no attribution modules or scalar game.}
\label{tab:close-work-map}
\end{table*}

\paragraph{Reading the map.}
Population CCA/MRCCA \citep{budhathoki2021why,quintasmartinez2024multiply} is
the closest population-attribution framework: it
becomes the same coalition game when each function/law pair is treated as one
node player. Fixed-SCM outlier RCA \citep{budhathoki2022causal} shares a
factual-noise replacement orientation, but attributes a different event-level
tail functional. Unit-level change attribution \citep{budhathoki2022unit}
instead switches observed values, and do-Shapley \citep{witter2026exactly}
switches variable interventions within one SCM. These distinctions explain why the
first two relations merit explicit mappings below, while the remaining rows
are adjacent causal questions rather than reductions of the population
alarm-score estimand.

\paragraph{Paired-node population reduction.}
Suppose the endpoints are Markovian, have aligned variables, and share a
common acyclic parent universe.  If each pair
\(\{\mathtt F_j,\mathtt Q_j\}\) is switched as one player, every hybrid law
factorizes into the induced conditionals.  For a Borel set
\(A\subseteq\mathbb R\), define
\[
C_j^{(r)}(A\mid x_{<j})
:=Q_j^{(r)}\!\left(\left\{e_j:
F_j^{(r)}(x_{<j},e_j)\in A\right\}\right).
\]
Specifically,
\[
P^U(dx)=\prod_{j=1}^d
C_j^{\mathbf 1\{\{\mathtt F_j,\mathtt Q_j\}\in U\}}
(dx_j\mid x_{<j}).
\]
Thus, for \(T(M)=\mathbb E_M[\tau(X)]\), this paired S$^4$-RCA game agrees
coalition by coalition with node-wise CCA using outcome \(\tau(X)\)
\citep{budhathoki2021why,quintasmartinez2024multiply}, and so has the same
Shapley values.  This reduction does not imply
\(\Psi_j=\Phi_{\mathtt F_j}+\Phi_{\mathtt Q_j}\) in the separately ordered
flat game, because splitting a player redistributes interactions.

\paragraph{Fixed-SCM event mapping.}
The native fixed-SCM outlier game \citep{budhathoki2022causal} indexes a coalition \(I\) by the noises
randomized from \(\delta_{e^{(0)}(x)}\) back to \(Q^{(0)}\), and uses the
negative log tail probability.  To express the same estimand in this paper's
reference-to-event orientation, let \(S=[d]\setminus I\) index the noises
fixed to their factual values and set
\[
w(S)=\log P_{M(F^{(0)},\,\delta_{e_S^{(0)}(x)}
                     \otimes Q_{-S}^{(0)})}(A_x).
\]
If \(v(I)\) is the native negative-log randomization game, then
\(w(S)=-v([d]\setminus S)\); complementing coalitions and reversing the sign
leaves every Shapley value unchanged.  Thus Table~\ref{tab:close-work-map}
correctly writes its source and target as \(Q^{(0)}\) and
\(\delta_{e^{(0)}(x)}\); their \(T_x\) contrast equals
\(S^\tau_{M^{(0)}}(x)\).
Our fixed-SCM event game in Sec.~C.3 has this same replacement orientation
but a different, expected-score functional; the outlier formula and
coherence identity therefore do not transfer automatically.  The paired
reduction and event mapping delimit the prior-work boundary; function/law
resolution, exact second-order computation, and SCM-wide scarce-incident
attribution are the contributions studied here.

\paragraph{FANS and nonlinear function/law dissection.}
FANS shows that function/noise dissection need not be confined to additive
models \citep{seo2026fans}.  It considers the following nonlinear,
non-additive class, with a differentiable noise inverse for each fixed \(h\):
\[
\begin{aligned}
F_j^{(r)}(pa_j,e_j)
&=g_j\!\left(h_j^{(r)}(pa_j),e_j\right),\\[-2pt]
e_j&\mapsto g_j(h,e_j)
\quad\text{is a differentiable bijection}.
\end{aligned}
\]
where the link \(g_j\) is invariant across regimes.  Within this class, a
function shift is distinguishable from a law shift when
\(g_j(h_j^{(1)}(pa_j),e_j^{(0)})\) cannot be rewritten, simultaneously for
every \(pa_j\), as
\(g_j(h_j^{(0)}(pa_j),\psi(e_j^{(0)}))\) using one parent-independent
invertible noise map \(\psi\).  Under its pure-shift setup, such a function
shift makes the reference-inferred residual depend on the parents, whereas a
pure law shift preserves independence.  If the rewriting is possible, the two
explanations are observationally indistinguishable; simultaneous changes
require a further restriction in FANS.  FANS is therefore a complementary
upstream shift-type diagnostic: our game attributes the propagated score
effects of aligned components, while the exact computation and scarce-incident
guarantees specialize to ordered linear SCMs.

\subsection{Hierarchy-preserving Owen attribution}
\label{app:hierarchical-owen}

Let \(N=\{\mathtt F_j,\mathtt Q_j:j\in[d]\}\) be the component player set,
let \(v(A)=T(M^A)\), and equip \(N\) with the a priori-union partition
\(\mathcal P=\{P_1,\ldots,P_d\}\), where
\(P_j=\{\mathtt F_j,\mathtt Q_j\}\).
For \(R\subseteq[d]\), write \(P(R)=\bigcup_{k\in R}P_k\) and
\[
w_d(R)=\frac{|R|!(d-|R|-1)!}{d!}.
\]
The Owen value \citep{owen1977values} averages marginal contributions over orders in which the two players of each block remain contiguous.
Since every block here has size two, its allocations have the explicit form
\begin{align}
\Omega_{\mathtt F_j}
&=\frac12\sum_{R\subseteq[d]\setminus\{j\}}w_d(R)
\bigl[
v\{P(R)\cup\{\mathtt F_j\}\}-v\{P(R)\}
\notag\\
&\quad+v\{P(R)\cup P_j\}-v\{P(R)\cup\{\mathtt Q_j\}\}
\bigr],
\label{eq:owen-f-explicit}\\
\Omega_{\mathtt Q_j}
&=\frac12\sum_{R\subseteq[d]\setminus\{j\}}w_d(R)
\bigl[
v\{P(R)\cup\{\mathtt Q_j\}\}-v\{P(R)\}
\notag\\
&\quad+v\{P(R)\cup P_j\}-v\{P(R)\cup\{\mathtt F_j\}\}
\bigr].
\label{eq:owen-q-explicit}
\end{align}
Let \(v^{\mathcal P}(R)=v\{P(R)\}\) be the quotient game and let
\(\Psi_j\) be its Shapley value, i.e., the paired-node population
attribution in Sec.~\ref{app:close-work-positioning}.  Adding
\eqref{eq:owen-f-explicit} and \eqref{eq:owen-q-explicit} cancels the two
intermediate coalition values, so
\begin{align}
\Omega_{\mathtt F_j}+\Omega_{\mathtt Q_j}=\Psi_j,
\qquad
\sum_{j=1}^d(\Omega_{\mathtt F_j}+\Omega_{\mathtt Q_j})
=v(N)-v(\varnothing).
\label{eq:owen-group-coherence-app}
\end{align}

For exact computation in the ordered-linear/quadratic-score
specialization, give node \(j\) the four categorical states
\(rs\in\{00,10,01,11\}\), let \(H(\pi)\) be the product-mixture expected score, and evaluate it by the moment recursion.
On the outer path \(\pi_{00,j}=1-t\), \(\pi_{11,j}=t\), and
\(\pi_{10,j}=\pi_{01,j}=0\), set
\(g_j^{rs}(t)=\partial H/\partial\pi_{rs,j}\),
\(D_{\mathrm{pair},j}=g_j^{11}-g_j^{00}\), and
\(D_{\mathrm{cross},j}=g_j^{10}-g_j^{01}\).
Expanding over the other \(d-1\) independently switched blocks and applying the Shapley beta-integral identity gives
\begin{align}
\Omega_{\mathtt F_j}
&=\frac12\int_0^1
\{D_{\mathrm{pair},j}(t)+D_{\mathrm{cross},j}(t)\}\,dt,
\notag\\
\Omega_{\mathtt Q_j}
&=\frac12\int_0^1
\{D_{\mathrm{pair},j}(t)-D_{\mathrm{cross},j}(t)\}\,dt.
\label{eq:owen-pair-cross-integrals}
\end{align}
Each difference is a Bernstein polynomial of degree at most \(d-1\), so
\(\lceil d/2\rceil\)-point Gauss--Legendre quadrature is exact.
One reverse sweep returns all local categorical tangents in \(O(d^3)\) time
and \(O(d^2)\) memory per point.  Hence the full Owen vector costs
\(O(d^4)\) time and \(O(d^2)\) memory.

Linearity in coalition values also preserves event--population coherence for
these prespecified coarse \(F_j/Q_j\) players.

\FloatBarrier
\section{Algorithms and Implementation Details}
\label{app:algorithms}

This section turns the computation and estimation results into two executable
procedures.  Algorithm~\ref{alg:exact-second-order-attribution} takes two SCM
endpoint moment sets and a fixed quadratic score, propagates the interpolated
moments in causal order, and obtains every resolved Shapley value through one
shared reverse sweep at each Gauss--Legendre point.  The reverse sweep
differentiates the moment program; it neither reverses causal edges nor defines
a reverse SCM.  Supplement Sec.~C.2 gives the corresponding recursions and
proof.  Each quadrature point costs \(O(d^3)\) dense time and \(O(d^2)\) memory,
and the resolved \(K=2d\) game uses \(d\) points.

Algorithm~\ref{alg:cross-fitted-plugin} implements the scarce-incident
estimator.  It fits the reference endpoint once, learns incident deviations
only on training folds, evaluates residual moments on held-out rows, and sends
each fold-matched endpoint pair to the exact routine before averaging.
Supplement Secs.~C.4--C.7 give the estimation conditions and guarantees.

\begin{algorithm}[!t]
\caption{Exact second-order attribution for the resolved function/law game}
\label{alg:exact-second-order-attribution}
\begin{algorithmic}[1]
\INPUT Endpoint moments \(\{b_j^{(r)},\mu_j^{(r)},\omega_j^{(r)}\}_{j,r}\); symmetric \(W\); \(z\)
\OUTPUT Resolved attribution vector \(\Phi\)
\STATE \emph{Gauss--Legendre quadrature and endpoint setup:}
Set \(n_{\rm q}=d\) and the symmetric tridiagonal \(G\in\mathbb R^{n_{\rm q}\times n_{\rm q}}\) by
\(G_{kk}=0\) for \(k\in[n_{\rm q}]\) and
\(G_{k,k+1}=G_{k+1,k}=k/\sqrt{4k^2-1}\) for \(k\in[n_{\rm q}-1]\)
\STATE Compute \(G=V\operatorname{Diag}(\xi_1,\ldots,\xi_{n_{\rm q}})V^\top\)
with orthonormal eigenvectors, then set
\(t_\ell\leftarrow(1+\xi_\ell)/2\), \(w_\ell\leftarrow V_{1\ell}^2\)
\STATE Initialize \(\Phi\leftarrow 0\)
\FOR{\(j=1,\ldots,d\)}
  \STATE \(\Delta b_j\leftarrow b_j^{(1)}-b_j^{(0)}\),
  \(\Delta\mu_j\leftarrow\mu_j^{(1)}-\mu_j^{(0)}\)
  \STATE \(\Delta\Gamma_j\leftarrow b_j^{(1)}(b_j^{(1)})^\top-b_j^{(0)}(b_j^{(0)})^\top\),
  \(\Delta q_j\leftarrow(\mu_j^{(1)})^2+\omega_j^{(1)}-(\mu_j^{(0)})^2-\omega_j^{(0)}\)
\ENDFOR
\STATE \emph{Forward--reverse evaluation at quadrature nodes:}
\FOR{\(\ell=1,\ldots,n_{\rm q}\)}
  \STATE \(t\leftarrow t_\ell\), \(p\leftarrow t\mathbf 1\)
  \FOR{\(j=1,\ldots,d\)}
    \STATE \(\bar b_j\leftarrow(1-t)b_j^{(0)}+t b_j^{(1)}\),
    \(\bar\mu_j\leftarrow(1-t)\mu_j^{(0)}+t\mu_j^{(1)}\)
    \STATE \(\Gamma_j\leftarrow(1-t)b_j^{(0)}(b_j^{(0)})^\top+t b_j^{(1)}(b_j^{(1)})^\top\)
    \STATE \(q_j\leftarrow(1-t)\{(\mu_j^{(0)})^2+\omega_j^{(0)}\}
    +t\{(\mu_j^{(1)})^2+\omega_j^{(1)}\}\), \(c_j\leftarrow\bar b_j\bar\mu_j\)
  \ENDFOR
  \STATE Run the forward causal-moment recursion in~\eqref{eq:main-moment-recursion}
  to obtain \(\bar x(p)\) and \(S(p)\)
  \STATE Initialize the shared reverse sweep with sensitivities
  \(a\leftarrow z\) and \(H\leftarrow W\)
  \FOR{\(j=d,d-1,\ldots,1\)}
    \STATE \(\alpha_j\leftarrow a_j\),
    \(\rho_j\leftarrow H_{<j,j}+H_{j,<j}^\top\),
    \(\gamma_j\leftarrow H_{jj}\)
    \STATE \(g_{\mathtt F_j}\leftarrow
    (\alpha_j\bar x_{<j}+S_{<j,<j}^\top\rho_j
    +2\gamma_j\bar\mu_j\bar x_{<j})^\top\Delta b_j
    +\gamma_j\langle\Delta\Gamma_j,S_{<j,<j}\rangle_{\rm F}\)
    \STATE \(g_{\mathtt Q_j}\leftarrow
    (\alpha_j+\rho_j^\top\bar x_{<j}
    +2\gamma_j\bar b_j^\top\bar x_{<j})\Delta\mu_j
    +\gamma_j\Delta q_j\)
    \STATE \(a_{<j}\mathrel{+}=\alpha_j\bar b_j+\rho_j\bar\mu_j
    +2\gamma_j\bar\mu_j\bar b_j\)
    \STATE \(H_{<j,<j}\mathrel{+}=\rho_j\bar b_j^\top+\gamma_j\Gamma_j\)
  \ENDFOR
  \STATE Accumulate \(\Phi\mathrel{+}=w_\ell g\)
\ENDFOR
\STATE return \(\Phi\)
\end{algorithmic}
\end{algorithm}

\begin{algorithm}[!t]
\caption{Sample-efficient reference-anchored cross-fitted plug-in attribution}
\label{alg:cross-fitted-plugin}
\begin{algorithmic}[1]
\INPUT Reference data \(\mathcal D^{(0)}\); incident data \(\mathcal D^{(1)}\);
causal order; prespecified split and tuning rules; optional \(W,z\)
\OUTPUT Fold-averaged attribution \(\overline{\widehat\Phi}\)
\STATE Split reference rows into fit and moment sets
\((\mathcal D_{\rm fit}^{(0)},I_0)\); fit preprocessing and optional \(W,z\)
on \(\mathcal D_{\rm fit}^{(0)}\) and apply the map to both regimes
\FOR{\(k=1,\ldots,d\)}
  \STATE Tune \(\lambda_k^{(0)}\) on reference-fit rows and set
  \(\widehat b_k^{(0)}\in\arg\min_b
  \{|\mathcal D^{(0)}_{\rm fit}|^{-1}\|X_{\rm fit,k}^{(0),c}
  -X_{\rm fit,<k}^{(0),c}b\|_2^2/2+\lambda_k^{(0)}\|b\|_1\}\)
\ENDFOR
\STATE On \(I_0\), compute \(\widehat e_i^{(0)}
=(I_d-\widehat B^{(0)\top})\widetilde x_i^{(0)}\) and its coordinatewise
mean/variance \((\widehat\mu^{(0)},\widehat\omega^{(0)})\)
\STATE Form incident holdout folds \(I_1,\ldots,I_L\) with
\(J_\ell=[n_1]\setminus I_\ell\)
\FOR{\(\ell=1,\ldots,L\)}
  \FOR{\(k=1,\ldots,d\)}
    \STATE \(y_{\ell k}\leftarrow X_{\ell,J_\ell,k}^{(1),c}
    -X_{\ell,J_\ell,<k}^{(1),c}\widehat b_k^{(0)}\)
    \STATE Tune \(\lambda_{\ell k}\) on \(J_\ell\), fit
    \(\widehat\delta_{\ell k}\in\arg\min_\delta
    \{|J_\ell|^{-1}\|y_{\ell k}-X_{\ell,J_\ell,<k}^{(1),c}\delta\|_2^2/2
    +\lambda_{\ell k}\|\delta\|_1\}\), and set
    \(\widehat b_{\ell k}^{(1)}\leftarrow\widehat b_k^{(0)}+\widehat\delta_{\ell k}\)
  \ENDFOR
  \STATE On \(I_\ell\), compute residuals under \(\widehat B_\ell^{(1)}\)
  and their coordinatewise mean/variance
  \((\widehat\mu_\ell^{(1)},\widehat\omega_\ell^{(1)})\)
  \STATE Apply Algorithm~\ref{alg:exact-second-order-attribution} to the
  reference and fold-\(\ell\) endpoints, obtaining \(\widehat\Phi_\ell\)
\ENDFOR
\STATE return \(L^{-1}\sum_{\ell=1}^L\widehat\Phi_\ell\)
\end{algorithmic}
\end{algorithm}

Preprocessing is fitted on reference rows and applied unchanged to both
regimes.  Centering or intercepts may be used within training folds, but
held-out residual location remains part of \(Q\), and no intercept is counted
again as an \(F\) player; regime-specific recentering would erase this location
shift.
Out-of-fold residuals are never pooled into a game formed with a different
mechanism estimate.
If the reference regime is cross-fitted, each incident fold is analogously paired with its own reference endpoint.
For a learned rather than supplied monitor, a literal application of the theorem
uses a separate reference score-fitting split or requires joint control of score estimation.
The theorem uses a penalty satisfying the score event; experiments tune inside each training fold only.

\FloatBarrier
\section{Proofs and Technical Lemmas}
\label{app:proofs}

This appendix collects the proofs used by the main paper.
The order follows the reader's pipeline: efficiency and coherence, exact computation,
its fixed-SCM event specialization, high-dimensional definitions, Delta-Lasso,
held-out residual moments, and finally plug-in attribution and ranking.

\subsection{Proof of Proposition~\ref{prop:direct-population-coherence}}
\label{app:proof-state-replacement-efficiency}
\label{app:proof-direct-population-coherence}

\begin{proof}[Proof of Proposition~\ref{prop:direct-population-coherence}]
By construction, \(M^\emptyset=M^{(0)}\); by the grand-coalition assumption,
\(M^{\mathcal U}=M^{(1)}\).  Finite-game
Shapley efficiency applied to \eqref{eq:causal-shapley} gives
\[
\sum_{u\in\mathcal U}\Phi_u
=T(M^{(1)})-T(M^{(0)}).
\]

For coherence, fix a coalition \(U\subseteq\mathcal U\) and draw
\(e\sim Q^{(1)}\).  Let \(M^U_{\delta_e}\) be the hybrid whose target is
\(M(F^{(1)},\delta_e)\), and let \(M^U_{Q^{(1)}}\) be the corresponding
population hybrid whose target is \(M(F^{(1)},Q^{(1)})\).  The two hybrids have
the same structural functions.  Averaging the selected point-mass
coordinates in \(M^U_{\delta_e}\) over \(e\sim Q^{(1)}\) gives their target
laws \(Q_j^{(1)}\), while every unselected coordinate retains its source law
\(Q_j^{(0)}\).  Independence of the exogenous coordinates therefore yields the
coalition-wise identity
\[
\mathbb E_{e\sim Q^{(1)}}\!\left[T(M^U_{\delta_e})\right]
=T(M^U_{Q^{(1)}}).
\]
Each Shapley value is a finite linear combination of these coalition values.
Taking expectations term by term in \eqref{eq:causal-shapley} proves
\(\mathbb E_{e\sim Q^{(1)}}[\Phi_u^{\mathrm{shift}}(e)]
=\Phi_u\) for every player \(u\).
\end{proof}

\subsection{Exact moment-recursion SCM-replacement Shapley: proof of Theorem~\ref{thm:population-moment-recursion}}
\label{app:proof-moment-recursion-shapley}

\begin{proof}[Proof of Theorem~\ref{thm:population-moment-recursion}]
Use the notation of the theorem.  For node \(j\), write \(b_j^{R_p}\)
and \(e_j^{R_p}\) for the coefficient vector and exogenous variable selected
by \(R_p\), and let
\(\mu_j^{R_p}=\mathbb E[e_j^{R_p}\mid R_p]\) and
\(\omega_j^{R_p}=\operatorname{Var}(e_j^{R_p}\mid R_p)\).
Conditional on \(R_p\), the ordered linear SCM satisfies
\[
x_j=(b_j^{R_p})^\top x_{<j}+e_j^{R_p}.
\]
Because modules are node-local and independently switched in the multilinear extension, the current node's switching indicators are independent of \(x_{<j}\), which depends only on earlier structural equations, earlier switches, and exogenous variables.
Therefore
\[
\mathbb E[x_j]
=
\mathbb E[b_j^{R_p}]^\top \mathbb E[x_{<j}]
+
\mathbb E[e_j^{R_p}]
=
\bar b_j^\top \bar x_{<j}+\bar\mu_j.
\]
For the raw cross second moment,
\[
\mathbb E[x_{<j}x_j]
=
\mathbb E[x_{<j}x_{<j}^\top]\bar b_j
+
\mathbb E[x_{<j}]\bar\mu_j,
\]
which is the second line of \eqref{eq:main-moment-recursion}.
For the diagonal raw second moment,
\begin{align*}
\mathbb E[x_j^2]
&=
\mathbb E\!\left[(b_j^{R_p})^\top x_{<j}x_{<j}^\top b_j^{R_p}\right]
\\
&\quad+
2\,\mathbb E\!\left[\mu_j^{R_p}(b_j^{R_p})^\top\right]\bar x_{<j}
+\mathbb E[(e_j^{R_p})^2]
\\
&=
\mathrm{tr}\!\left(\Gamma_jS_{<j,<j}\right)
+2c_j^\top \bar x_{<j}
+q_j.
\end{align*}
This proves the recursion in \eqref{eq:main-moment-recursion} for
\(\bar x(p)\) and \(S(p)\).
Since \(\tau(x)=x^\top W x+z^\top x\), the multilinear extension is
\begin{align*}
\widetilde T(p)
&=\mathbb E_{R_p}\!\left[T(M^{R_p})\right]
\\
&=\mathbb E[\tau(x)]
\\
&=
\mathrm{tr}(WS(p))+z^\top \bar x(p),
\end{align*}
which is \eqref{eq:main-random-switch-target}.

It remains to connect this program to the Shapley weights.  Differentiating
\eqref{eq:main-multilinear-extension} directly with respect to \(p_u\)
and pairing coalitions that differ only in player \(u\) gives
\begin{align*}
\frac{\partial\widetilde T(p)}{\partial p_u}
&=\sum_{U\subseteq\mathcal U\setminus\{u\}}
\left\{T(M^{U\cup\{u\}})-T(M^U)\right\}\\
&\quad\times
\prod_{v\in U}p_v
\prod_{v\in\mathcal U\setminus(U\cup\{u\})}(1-p_v).
\end{align*}
Consequently, with \(K=|\mathcal U|\),
\begin{align*}
\frac{\partial\widetilde T(t\mathbf 1)}{\partial p_u}
&=\sum_{U\subseteq\mathcal U\setminus\{u\}}
\left\{T(M^{U\cup\{u\}})-T(M^U)\right\}\\
&\quad\times t^{|U|}(1-t)^{K-|U|-1}.
\end{align*}
Integrating term by term over \(t\in[0,1]\) and using
\[
\int_0^1 t^{|U|}(1-t)^{K-|U|-1}\,dt
=
\frac{|U|!\,(K-|U|-1)!}{K!}
\]
recovers \eqref{eq:causal-shapley} and proves
\eqref{eq:main-shapley-multilinear-integral} directly
\citep{shapley1953value,owen1972multilinear}.
The displayed derivative is a polynomial in \(t\) of degree at most \(K-1\).
An \(m\)-point Gauss--Legendre rule is exact for polynomials of degree at most \(2m-1\), so \(m=\lceil K/2\rceil\) is sufficient; standard nodes and weights can be computed by the Golub--Welsch procedure \citep{golub1969calculation}.
\end{proof}

\begin{proof}[Proof of Corollary~\ref{cor:exact-all-component}]
We give the backward-sensitivity equations for the separated function/law game;
paired modules use the same sweep with the mixed local moment \(c_j\), and
Appendix~\ref{app:hierarchical-owen} gives the Owen categorical tangents.
Fix a probability vector \(p\) and run
\eqref{eq:main-moment-recursion} forward.
In reverse node order, let \(\alpha_j\), \(\rho_j\), and \(\gamma_j\) denote respectively the current sensitivity coefficients of \(\bar x_j\), of the single cross-moment vector after summing the two copied-triangle sensitivities, and of \(S_{jj}\).
For the separated game, write \(b_j=\bar b_j\), \(\mu_j=\bar\mu_j\),
\(\Gamma_j=\mathbb E[b_j^{R_p}(b_j^{R_p})^\top]\), and define endpoint differences
\begin{align*}
\Delta b_j&=b_j^{(1)}-b_j^{(0)},
\qquad \Delta\mu_j=\mu_j^{(1)}-\mu_j^{(0)},\\
\Delta\Gamma_j&=b_j^{(1)}(b_j^{(1)})^\top-b_j^{(0)}(b_j^{(0)})^\top,\\
q_j^{(r)}&=(\mu_j^{(r)})^2+\omega_j^{(r)},
\qquad \Delta q_j=q_j^{(1)}-q_j^{(0)}.
\end{align*}
Differentiating the three forward assignments at node \(j\) gives both component derivatives at once:
\begin{align}
\frac{\partial\widetilde T}{\partial p_{\mathtt F_j}}
&=\left(\alpha_j\bar x_{<j}+S_{<j,<j}^\top\rho_j
+2\gamma_j\mu_j\bar x_{<j}\right)^\top\Delta b_j
\notag\\
&\quad+\gamma_j\langle\Delta\Gamma_j,S_{<j,<j}\rangle,
\label{eq:reverse-grad-f}\\
\frac{\partial\widetilde T}{\partial p_{\mathtt Q_j}}
&=\left(\alpha_j+\rho_j^\top\bar x_{<j}
+2\gamma_j b_j^\top\bar x_{<j}\right)\Delta\mu_j
+\gamma_j\Delta q_j.
\label{eq:reverse-grad-q}
\end{align}
Writing \(a\) and \(H\) for the sensitivity arrays initialized by
\(a=z\) and \(H=W\), the updates passed to the already-computed prefix are
\begin{align}
a_{<j}
&\mathrel{+}=\alpha_jb_j+\rho_j\mu_j+2\gamma_j\mu_jb_j,
\notag\\
H_{<j,<j}
&\mathrel{+}=\rho_jb_j^\top+\gamma_j\Gamma_j.
\label{eq:reverse-prefix}
\end{align}
The two triangle contributions are combined in \(\rho_j\).
Equations~\eqref{eq:reverse-grad-f}--\eqref{eq:reverse-prefix} apply the chain rule to the exact moment program, so they return all \(2d\) derivatives without approximation.

The forward and backward matrix--vector/inner-product work summed over nodes is \(O(d^3)\), and the mean, second-moment, and sensitivity arrays occupy \(O(d^2)\) memory.
There are \(\lceil K/2\rceil=O(K)\) exact quadrature points.
Hence the full vector costs \(O(Kd^3)\); setting \(K=2d\) gives \(O(d^4)\).
Evaluating every derivative through two one-player-at-a-time forward programs would instead cost \(O(K^2d^3)=O(d^5)\).
The paired game has \(K=d\), and the construction in
Appendix~\ref{app:hierarchical-owen} establishes the same order for the
hierarchy-preserving report.
\end{proof}

\subsection{Fixed-SCM event specialization}
\label{app:fixed-scm-event-specialization}

\begin{corollary}[Fixed-SCM event attribution]
\label{cor:fixed-scm-event-specialization}
Fix an ordered linear SCM \(M(B,Q)\) and the quadratic score in
\eqref{eq:tau}.  Let \(\mu=\mathbb E_Q[e]\),
\(\Omega=\Diag(\omega)\), \(A=(I-B)^{-1}\), \(V=AWA^\top\), and
\(\ell=Az\).  For an event \(x\), set \(e=(I-B^\top)x\).  The \(Q\)-only
replacement game from \(M(B,Q)\) to \(M(B,\delta_e)\) has Shapley vector
\begin{align}
\Phi^{\mathrm{fix}}(e)
&=\Diag(e-\mu)\{V(e+\mu)+\ell\}-\diag(V\Omega).
\label{eq:fixed-scm-event-vector}
\end{align}
It satisfies
\(\mathbf 1^\top\Phi^{\mathrm{fix}}(e)
=\tau(x)-\mathbb E_{X\sim M(B,Q)}[\tau(X)]\).
Embedding the endpoints in the primary function/law-resolved player set adds only null \(F\)
players.
\end{corollary}

\begin{proof}
This is the fixed-mechanism specialization of
Theorem~\ref{thm:population-moment-recursion}.  Write
\(\Delta=e-\mu\).  Under independent coordinate switches, the switched
residual at coordinate \(j\) has first and second moments
\begin{align*}
m_j(p)&=\mu_j+p_j\Delta_j,\\
q_j(p)&=\mu_j^2+\omega_j
+p_j(e_j^2-\mu_j^2-\omega_j).
\end{align*}
Independence gives cross moments \(m_i(p)m_j(p)\) for \(i\ne j\), so the
multilinear extension computed by the theorem is
\[
\widetilde T(p)
=\sum_jV_{jj}q_j(p)
+\sum_{i\ne j}V_{ij}m_i(p)m_j(p)+\ell^\top m(p).
\]
Using \(V=V^\top\), differentiation at \(p=t\mathbf 1\) gives
\begin{align*}
\frac{\partial\widetilde T(t\mathbf 1)}{\partial p_j}
&=V_{jj}(e_j^2-\mu_j^2-\omega_j)\\
&\quad+2\Delta_j\sum_{k\ne j}V_{jk}(\mu_k+t\Delta_k)
+\ell_j\Delta_j.
\end{align*}
Integration over \(t\in[0,1]\), as prescribed by the theorem, yields
\[
(\Phi^{\mathrm{fix}})_j
=\Delta_j\{V(e+\mu)+\ell\}_j-V_{jj}\omega_j,
\]
which is \eqref{eq:fixed-scm-event-vector} because
\(\diag(V\Omega)_j=V_{jj}\omega_j\).  The efficiency identity follows
from finite-game Shapley efficiency; the null-player
claim follows because the mechanism is identical at both endpoints.
\end{proof}

\subsection{High-dimensional conditions and auxiliary estimators}

\paragraph{Identification versus estimation.}
The nonsingular parent covariance invoked in the main-text identification
paragraph is a population condition, not invertibility of an empirical
high-dimensional covariance.  With known parent sets, it is needed only on
the parent subvector.  With only a known order and a parameter class of
nodewise sparsity at most \(s\ge1\), it can be weakened to restricted uniqueness:
\[
\inf_{\substack{\delta\ne0\\\|\delta\|_0\le2s}}
\frac{\delta^\top\operatorname{Cov}(x_{<j}^{(r)})\delta}
{\|\delta\|_2^2}>0
\quad\text{for every \(r\) and \(j>1\)}.
\]
Indeed, the difference of two \(s\)-sparse population slopes is
\(2s\)-sparse.  This identification condition is logically distinct from
the sample RE/RSC assumptions below, which control estimation error.

\paragraph{Notation.}
Nodewise slope regressions use centered or intercept-residualized designs,
denoted by \(\mathbf X^{(r)}\).  For node \(k\), let
\(\mathbf Z_k^{(r)}=\mathbf X_{:,1:(k-1)}^{(r)}\),
\(S_k^{(r)}=\operatorname{supp}(b_k^{(r)})\),
\(s_{r,k}=|S_k^{(r)}|\), and \(s_r=\sum_ks_{r,k}=\|B^{(r)}\|_0\).
For \(S\subseteq[k-1]\), throughout this appendix
\(S^c=[k-1]\setminus S\) is the complement relative to the coordinates of
the node-\(k\) regression.  For matrices,
\(\|M\|_{1,1}=\sum_{i,j}|M_{ij}|\), and
\(\Delta=B^{(1)}-B^{(0)}\).

\paragraph{Sub-Gaussian noise, residualization, and the joint score event.}
For a scalar random variable \(Y\) and random vector \(X\), define
\begin{align*}
\|Y\|_{\psi_2}
&:=\inf\{c>0:\mathbb E\exp(Y^2/c^2)\le2\},\\
\|X\|_{\psi_2}
&:=\sup_{\|v\|_2=1}\|v^\top X\|_{\psi_2}.
\end{align*}
Write
\(\boldsymbol\varepsilon_k^{(r)}
=(\varepsilon_{1k}^{(r)},\ldots,\varepsilon_{n_rk}^{(r)})^\top\)
for the centered row disturbances \(e_k^{(r)}-\mu_k^{(r)}\), before sample
centering or intercept residualization.  Let \(P^{(r)}\) be the corresponding deterministic
symmetric idempotent residualizer, with \(P^{(r)}\mathbf 1=0\), and
\(\mathbf Z_k^{(r)}=P^{(r)}\mathbf Z_{k,\mathrm{raw}}^{(r)}\).
Conditional on the raw parent design, assume the
coordinates of \(\boldsymbol\varepsilon_k^{(r)}\) are independent, centered,
and, for all \(i\in[n_r]\), \(k\in[d]\), and \(t\in\mathbb R\),
\begin{align}
\mathbb E\!\left[\exp\{t\varepsilon_{ik}^{(r)}\}
\mid\mathbf Z_{k,\mathrm{raw}}^{(r)}\right]
\le\exp(\sigma_r^2t^2/2).
\label{eq:conditional-subgaussian-definition}
\end{align}
Column normalization means
\begin{align}
\max_{k\in[d]}\max_{j<k}
\frac{\|\mathbf Z_{k,:,j}^{(r)}\|_2}{\sqrt{n_r}}
\le1.
\label{eq:column-normalization-definition}
\end{align}
The residualized error
\(\bar{\boldsymbol\varepsilon}_k^{(r)}
:=P^{(r)}\boldsymbol\varepsilon_k^{(r)}\) need not have independent
coordinates.  Independence is not imposed after residualization: symmetry
and idempotence give
\[
\mathbf Z_k^{(r)\top}\bar{\boldsymbol\varepsilon}_k^{(r)}
=\mathbf Z_k^{(r)\top}\boldsymbol\varepsilon_k^{(r)}.
\]
For regime \(r\), choose
\begin{align}
\lambda_r
&=2\sigma_r\sqrt{\frac{2\log(d^2/\delta_r)}{n_r}},
\label{eq:joint-nodewise-lambda}
\end{align}
or any larger order-equivalent value, where \(0<\delta_r<1\).
Under~\eqref{eq:conditional-subgaussian-definition}--\eqref{eq:column-normalization-definition},
conditional concentration followed by a union bound gives the joint score event
\begin{align}
\max_{k\in[d]}\max_{j<k}
\left|n_r^{-1}\mathbf Z_{k,:,j}^{(r)\top}
\boldsymbol\varepsilon_k^{(r)}\right|
&\le \lambda_r/2
\label{eq:joint-nodewise-score-event}
\end{align}
with probability at least \(1-\delta_r\).
In the high-dimensional regime, any \(\delta_r\to0\) with
\(\log(1/\delta_r)=O(\log d)\) preserves
\(\lambda_r\asymp\sigma_r\sqrt{\log d/n_r}\) while controlling all nodewise
regressions with probability tending to one.

\paragraph{Restricted curvature.}
For \(S\subseteq[k-1]\), define the Lasso cone
\begin{align}
\mathcal C(S;3)
&:=\{\theta\in\mathbb R^{k-1}:
\|\theta_{S^c}\|_1\le3\|\theta_S\|_1\}.
\label{eq:lasso-cone-definition}
\end{align}
The cone restricted-eigenvalue condition used for sparse reference and
incident-only fits is \citep{bickel2009simultaneous}
\begin{align}
\frac{1}{n}\|\mathbf Z\theta\|_2^2
&\ge \kappa\|\theta\|_2^2
\quad\text{for every }\theta\in\mathcal C(S;3).
\label{eq:mca-re1}
\end{align}
Our \(\kappa\) absorbs the square in conventions that state the lower bound
for \(\|\mathbf Z\theta\|_2/\sqrt n\).  The incident transfer proof instead
uses the quadratic-loss specialization of restricted strong convexity
\citep{negahban2012unified}:
\begin{align}
\frac{1}{n}\|\mathbf Z\theta\|_2^2
&\ge \kappa_1\|\theta\|_2^2
-\kappa_2\frac{\log d}{n}\|\theta\|_1^2
\quad\text{for every }\theta\in\mathbb R^{k-1}.
\label{eq:mca-re2}
\end{align}
We call \eqref{eq:mca-re1} RE and \eqref{eq:mca-re2} RSC below.

These conditions are assumptions, not consequences of SCM linearity.
They hold with high probability for Gaussian random designs under
well-conditioned population covariance, typically once
\(n\gtrsim s\log d\); bounded-away-from-zero population eigenvalues yield
the global RSC form \citep{raskutti2010restricted,negahban2012unified}.
A union bound over \(O(d^2)\) node--coordinate pairs
retains the \(\log d\) factor.  Temporal dependence, collinearity, or
vanishing population curvature is outside the theorem.

\paragraph{Auxiliary Lasso estimators.}
The pre-incident reference estimator used by Delta-Lasso is
\begin{align}
\widetilde b_k
\in
\argmin_b
\left\{
\frac{1}{2n_0}
\|\mathbf X^{(0)}_{:,k}-\mathbf Z_k^{(0)}b\|_2^2
+\lambda_0\|b\|_1
\right\}.
\label{eq:mca-ref}
\end{align}
For the incident-only comparison in
Lemma~\ref{prop:incident-only-lasso-bound}, define
\begin{align}
\widehat b^{\mathrm{naive}}_k
\in
\argmin_b
\left\{
\frac{1}{2n_1}
\|\mathbf X^{(1)}_{:,k}-\mathbf Z_k^{(1)}b\|_2^2
+\lambda_1\|b\|_1
\right\}.
\label{eq:mca-naive}
\end{align}

\subsection{Reference, Delta-Lasso, and incident-only bounds}
\label{app:proof-mca-delta}

\begin{lemma}[Standard nodewise Lasso bounds]
\label{lem:standard-nodewise-lasso-bounds}
\label{prop:reference-lasso-bound}
\label{prop:incident-only-lasso-bound}
Under the joint score event and the nodewise RE condition,
\begin{align}
\|\widetilde B-B^{(0)}\|_{1,1}
&\le \frac{12\lambda_0s_0}{\kappa_0},
\label{eq:mca-ref-l1}\\
\|\widetilde B-B^{(0)}\|_{\mathrm F}^2
&\le \frac{9\lambda_0^2s_0}{\kappa_0^2},
\label{eq:mca-ref-frobenius}\\
\|\widehat B^{\mathrm{naive}}-B^{(1)}\|_{\mathrm F}^2
&\le \frac{9\lambda_1^2s_1}{\kappa^2}.
\label{eq:mca-naive-frobenius}
\end{align}
With \eqref{eq:joint-nodewise-lambda}, the reference and incident bounds
hold with probabilities at least \(1-\delta_0\) and \(1-\delta_1\),
respectively.
\end{lemma}

\begin{proof}
We give the common argument to make the constants and simultaneous event
explicit.  For either sparse endpoint, suppress \(r,k\), let
\(\widehat h=\widehat b-b\), \(S=\operatorname{supp}(b)\), and write the
centered regression as \(y=Zb+\bar\varepsilon\).  Comparing the Lasso
objective at \(\widehat b\) and \(b\), expanding the square, and using
\(\|Z^\top\bar\varepsilon/n\|_\infty\le\lambda/2\) yields
\citep{bickel2009simultaneous,negahban2012unified}
\begin{align}
\frac{\|Z\widehat h\|_2^2}{2n}
+\frac{\lambda}{2}\|\widehat h_{S^c}\|_1
&\le\frac{3\lambda}{2}\|\widehat h_S\|_1.
\label{eq:reference-basic-inequality}
\end{align}
Hence \(\widehat h\in\mathcal C(S;3)\).  The RE bound and
\(\|\widehat h_S\|_1\le\sqrt{|S|}\|\widehat h\|_2\) give
\[
\|\widehat h\|_2
\le\frac{3\lambda\sqrt{|S|}}{\kappa},
\qquad
\|\widehat h\|_1
\le4\|\widehat h_S\|_1
\le\frac{12\lambda|S|}{\kappa}.
\]
Applying these inequalities with
\((r,\lambda,\kappa,S)=(0,\lambda_0,\kappa_0,S_k^{(0)})\) and summing the
\(\ell_1\) and squared \(\ell_2\) bounds over \(k\) proves
\eqref{eq:mca-ref-l1}--\eqref{eq:mca-ref-frobenius}.
Applying the same argument with
\((1,\lambda_1,\kappa,S_k^{(1)})\) and summing squared \(\ell_2\) bounds
proves \eqref{eq:mca-naive-frobenius}.
The score event in~\eqref{eq:joint-nodewise-score-event} is already
simultaneous over all node--coordinate pairs, so no additional nodewise
union bound is needed.
\end{proof}

Recall from \eqref{eq:main-delta-radius} that
\[
R_\Delta
=\|\Delta\|_{1,1}+\frac{12\lambda_0s_0}{\kappa_0}.
\]
The tolerance condition in Theorem~\ref{thm:mca-delta-core} is
\(\kappa_2R_\Delta\sqrt{\log d/n_1}\le c\sigma_1\).
On the event in Lemma~\ref{prop:reference-lasso-bound}, the triangle
inequality gives
\begin{align}
\|B^{(1)}-\widetilde B\|_{1,1}
&\le \|B^{(1)}-B^{(0)}\|_{1,1}
\notag\\
&\quad+\|\widetilde B-B^{(0)}\|_{1,1}
\le R_\Delta,
\label{eq:mca-radius}
\end{align}

\begin{proof}[Proof of Theorem~\ref{thm:mca-delta-core}]
Work on the reference event above and condition on the independent fit
\(\widetilde B\).  For node \(k\), set
\(\widetilde\Delta_k=b_k^{(1)}-\widetilde b_k\) and
\(\widehat\Delta_k=\widehat\delta_k-\widetilde\Delta_k\).
The transformed incident response in \eqref{eq:mca-delta-reparam}
therefore obeys
\(\mathbf y_k=\mathbf Z_k^{(1)}\widetilde\Delta_k
+\bar{\boldsymbol\varepsilon}_k^{(1)}\).  The Delta-Lasso objective at
\(\widehat\delta_k\) is no larger than at its true parameter
\(\widetilde\Delta_k\):
\begin{align*}
&\frac{1}{2n_1}
\|\bar{\boldsymbol\varepsilon}_k^{(1)}-\mathbf Z_k^{(1)}\widehat\Delta_k\|_2^2
+\lambda_1\|\widehat\delta_k\|_1\\
&\qquad\le
\frac{1}{2n_1}\|\bar{\boldsymbol\varepsilon}_k^{(1)}\|_2^2
+\lambda_1\|\widetilde\Delta_k\|_1.
\end{align*}
After expanding and cancelling the noise norm,
\begin{align*}
\frac{1}{2n_1}
\|\mathbf Z_k^{(1)}\widehat\Delta_k\|_2^2
{}+\lambda_1\|\widehat\delta_k\|_1
&\le
\frac{1}{n_1}\bar{\boldsymbol\varepsilon}_k^{(1)\top}
\mathbf Z_k^{(1)}\widehat\Delta_k\\
&\quad+\lambda_1\|\widetilde\Delta_k\|_1.
\end{align*}
The incident joint score event is the corresponding simultaneous event
obtained by the union bound in C.4.  Hence
\[
\frac{1}{n_1}\bar{\boldsymbol\varepsilon}_k^{(1)\top}
\mathbf Z_k^{(1)}\widehat\Delta_k
\le\frac{\lambda_1}{2}\|\widehat\Delta_k\|_1.
\]
Together with
\(\|\widehat\Delta_k\|_1
\le\|\widehat\delta_k\|_1+\|\widetilde\Delta_k\|_1\), this gives
\begin{align}
\frac{1}{2n_1}\|\mathbf Z_k^{(1)}\widehat\Delta_k\|_2^2
+\frac{\lambda_1}{2}\|\widehat\delta_k\|_1
&\le\frac{3\lambda_1}{2}\|\widetilde\Delta_k\|_1.
\label{eq:delta-basic-inequality}
\end{align}
In particular,
\begin{align}
\|\widehat\Delta_k\|_1&\le4\|\widetilde\Delta_k\|_1,
&
\frac{1}{n_1}\|\mathbf Z_k^{(1)}\widehat\Delta_k\|_2^2
&\le3\lambda_1\|\widetilde\Delta_k\|_1.
\label{eq:delta-radius-prediction}
\end{align}
Applying the incident RSC condition \eqref{eq:mca-re2}, as in the standard
restricted-convexity analysis of regularized estimators
\citep{negahban2012unified}, gives
\begin{align}
\kappa_1\|\widehat\Delta_k\|_2^2
&\le3\lambda_1\|\widetilde\Delta_k\|_1
+16\kappa_2\|\widetilde\Delta_k\|_1^2\frac{\log d}{n_1}.
\label{eq:delta-rsc-bound}
\end{align}
Because \(\widehat\Delta_k\) is the \(k\)th column of
\(\widehat B^\delta-B^{(1)}\), summing over \(k\), using
\(\sum_k\|\widetilde\Delta_k\|_1\le R_\Delta\) and
\(\sum_k\|\widetilde\Delta_k\|_1^2\le R_\Delta^2\), gives the unabsorbed
bound
\[
\|\widehat B^\delta-B^{(1)}\|_{\mathrm F}^2
\le\frac{3\lambda_1R_\Delta
+16\kappa_2R_\Delta^2\log d/n_1}{\kappa_1}.
\]
The tolerance condition and
\(\lambda_1\asymp\sigma_1\sqrt{\log d/n_1}\) make its second numerator term
at most \(C\lambda_1R_\Delta\), proving
\eqref{eq:mca-delta-core-bound}.  The reference and incident
joint score events each already hold uniformly over all nodewise
regressions by the union bounds in C.4.  A final union bound across the two
regimes shows that they hold simultaneously with probability at least
\(1-\delta_0-\delta_1\), proving the stated high-probability claim.
\end{proof}

The tolerance condition is implied by
\(n_1\gtrsim(\kappa_2R_\Delta/\sigma_1)^2\log d\).
The argument controls the total \(\ell_1\) radius; true-shift sparsity alone
does not yield a faster rate for a reference-centered estimator with
estimated reference error.

\subsection{Held-out residual-moment estimation}
\label{app:proof-residual-moment-estimation}

We first bound the held-out residual mean and variance for one fitted regime,
then substitute the mechanism-estimation rates used in the main theorem.
Fold independence is a sampling condition of the out-of-sample construction,
not an additional moment restriction: conditional on the mechanism estimate,
the evaluation rows are independent and follow the corresponding regime law.

\begin{proposition}[Residual mean and variance estimation]
\label{prop:residual-moment-estimation}
For regime \(r\), put
\(\chi_r=\|A^{(r)}\|_{\mathrm{op}}\sigma_r\) and
\(M_r=\|\mathbb E x^{(r)}\|_2\).
Suppose a residual-evaluation fold of size \(m_r\) from regime \(r\) is
independent of its fitted mechanism and contains independent observations.
Assume the ordered SCM, the regime-\(r\) sub-Gaussian disturbance condition
of Sec.~C.4, and Assumption~\ref{ass:residual-moments}.
With \(\bar\eta_B=\max_j\|\widehat b_j-b_j\|_2\) and
\(u_{r,\zeta}=\sqrt{\log(4d/\zeta)/m_r}\), for every \(0<\zeta<1\), conditional on
the fitted mechanism and with probability at least \(1-\zeta\),
\begin{align}
\|\widehat\mu-\mu\|_\infty
&\le C\{\sigma_ru_{r,\zeta}
+(M_r+\chi_ru_{r,\zeta})\bar\eta_B\},
\label{eq:residual-mean-inf-bound}\\
\|\widehat\omega-\omega\|_\infty
&\le C\bigl\{\sigma_r^2(u_{r,\zeta}+u_{r,\zeta}^2)
+\sigma_r\chi_r\bar\eta_B(u_{r,\zeta}+u_{r,\zeta}^2)\notag\\
&\qquad\quad+\chi_r^2\bar\eta_B^2
(1+u_{r,\zeta}+u_{r,\zeta}^2)\bigr\}.
\label{eq:omega-inf-bound}
\end{align}
Here and below, \(C>0\) is a universal constant.
\end{proposition}

\begin{proof}
Condition on \(\widehat B\), fix a node, and write
\(h_j=\widehat b_j-b_j\),
\(\xi_i=x_i-\mathbb Ex\),
\(x_{i,<j}=\mathbb Ex_{<j}+\xi_{i,<j}\), and
\(e_{ij}=\mu_j+\varepsilon_{ij}\).  Then
the sub-Gaussian disturbance condition gives
\(\|\varepsilon_{ij}\|_{\psi_2}\le C\sigma_r\).
Since \(x-\mathbb Ex=A^{(r)\top}(e-\mu)\) and the exogenous coordinates are
independent,
\begin{align*}
\max_j\|\xi_{i,<j}\|_{\psi_2}
&\le C\chi_r,
&
\max_j\|\mathbb Ex_{<j}\|_2&\le M_r.
\end{align*}
Moreover,
\begin{align*}
\widehat e_{ij}
&=\mu_j-h_j^\top\mathbb Ex_{<j}+\varepsilon_{ij}+q_{ij},
&q_{ij}&=-h_j^\top\xi_{i,<j},
\end{align*}
where \(\|q_{ij}\|_{\psi_2}\le C\chi_r\|h_j\|_2\).
Consequently,
\[
\widehat\mu_j-\mu_j
=-h_j^\top\mathbb Ex_{<j}+\bar\varepsilon_j+\bar q_j.
\]
Sub-Gaussian concentration gives, simultaneously over \(j\),
\(|\bar\varepsilon_j|\le C\sigma_ru_{r,\zeta}\) and
\(|\bar q_j|\le C\chi_r\|h_j\|_2u_{r,\zeta}\).
Combining these inequalities with
\(|h_j^\top\mathbb Ex_{<j}|\le M_r\|h_j\|_2\), and taking a union bound
over the nodes, proves \eqref{eq:residual-mean-inf-bound}.

The constant term \(-h_j^\top\mathbb Ex_{<j}\) disappears when the fitted
residuals are centered.  Hence the variance error has the exact decomposition
\begin{align*}
\widehat\omega_j-\omega_j
&=m_r^{-1}\sum_i(\varepsilon_{ij}^2-\omega_j)
+2m_r^{-1}\sum_i\varepsilon_{ij}q_{ij}\\
&\quad+m_r^{-1}\sum_iq_{ij}^2-(\bar\varepsilon_j+\bar q_j)^2.
\end{align*}
Exogenous independence gives
\(\mathbb E(\varepsilon_{ij}q_{ij})=0\).
Moreover,
\(\|\varepsilon_{ij}^2\|_{\psi_1}\le C\sigma_r^2\),
\(\|\varepsilon_{ij}q_{ij}\|_{\psi_1}
\le C\sigma_r\chi_r\|h_j\|_2\), and
\(\|q_{ij}^2\|_{\psi_1}\le C\chi_r^2\|h_j\|_2^2\).
Bernstein's inequality therefore bounds the first two empirical averages by
\(C\sigma_r^2(u_{r,\zeta}+u_{r,\zeta}^2)\) and
\(C\sigma_r\chi_r\|h_j\|_2
\,(u_{r,\zeta}+u_{r,\zeta}^2)\), respectively.  For the third,
adding its expectation gives
\[
m_r^{-1}\sum_iq_{ij}^2
\le C\chi_r^2\|h_j\|_2^2
\,(1+u_{r,\zeta}+u_{r,\zeta}^2).
\]
Finally,
\((\bar\varepsilon_j+\bar q_j)^2
\le C(\sigma_r+\chi_r\|h_j\|_2)^2u_{r,\zeta}^2\).
A union bound over these events and all \(d\) nodes, followed by
\(\|h_j\|_2\le\bar\eta_B\), proves \eqref{eq:omega-inf-bound}.
\end{proof}

\begin{corollary}[Regimewise held-out moment rates]
\label{cor:regime-residual-moment-rates}
Define
\(\eta_B^{(r)}=\|\widehat B^{(r)}-B^{(r)}\|_{\mathrm F}\),
\(\eta_\mu^{(r)}=\|\widehat\mu^{(r)}-\mu^{(r)}\|_\infty\), and
\(\eta_\omega^{(r)}=\|\widehat\omega^{(r)}-\omega^{(r)}\|_\infty\).
Under Proposition~\ref{prop:residual-moment-estimation}, suppose
\(\eta_B^{(r)}=O_{\mathrm P}(\rho_r)\) and set
\(\nu_r=\sqrt{\log(2d)/m_r}\).  Then
\begin{align}
\eta_\mu^{(r)}
&=O_{\mathrm P}\!\left(\sigma_r\nu_r+(M_r+\chi_r\nu_r)\rho_r\right),
\label{eq:regime-residual-mean-rate}\\
\eta_\omega^{(r)}
&=O_{\mathrm P}\!\bigl(\sigma_r^2(\nu_r+\nu_r^2)
+\sigma_r\chi_r\rho_r(\nu_r+\nu_r^2)\notag\\
&\qquad\quad+\chi_r^2\rho_r^2(1+\nu_r+\nu_r^2)\bigr).
\label{eq:regime-residual-variance-rate}
\end{align}
Under Assumption~\ref{ass:residual-moments}, if \(\nu_r=o(1)\), these
reduce to
\(\eta_\mu^{(r)}=O_{\mathrm P}(\nu_r+(M_r+\nu_r)\rho_r)\) and
\(\eta_\omega^{(r)}=O_{\mathrm P}(\nu_r+\rho_r^2)\).
\end{corollary}

\begin{proof}
Every nodewise coefficient error is bounded by the matrix Frobenius error:
\(\max_j\|\widehat b_j^{(r)}-b_j^{(r)}\|_2\le\eta_B^{(r)}\).
For each fixed \(0<\zeta<1\),
\(\log(4d/\zeta)\le C_\zeta\log(2d)\), so
\(u_{r,\zeta}\le\sqrt{C_\zeta}\nu_r\).  This gives the two displayed
\(O_{\mathrm P}\) rates by taking \(\zeta\) arbitrarily small.
Assumption~\ref{ass:residual-moments} makes
\(\sigma_r,\chi_r=O(1)\).  When \(\nu_r=o(1)\), use
\(\rho_r\nu_r\le(\rho_r^2+\nu_r^2)/2\) and \(\nu_r^2\le\nu_r\)
and \(1+\nu_r+\nu_r^2=O(1)\) to obtain the simplified orders.
\end{proof}

\subsection{Plug-in attribution sensitivity and ranking}
\label{app:proof-population-plugin-rate}

The proof separates four reusable steps: hybrid-path stability, exact endpoint
gradients, pathwise norm envelopes, and deterministic rate sequences. Put
\begin{align*}
D_U^F&=\operatorname{Diag}
\bigl(\mathbf 1\{\mathtt F_j\in\bigcup_{w\in U}w\}\bigr)_{j=1}^d,\\
D_U^Q&=\operatorname{Diag}
\bigl(\mathbf 1\{\mathtt Q_j\in\bigcup_{w\in U}w\}\bigr)_{j=1}^d.
\end{align*}
For \(r\in\{0,1\}\), write
\(D_U^{F,(1)}=D_U^F\), \(D_U^{F,(0)}=I-D_U^F\), and analogously
\(D_U^{Q,(1)}=D_U^Q\), \(D_U^{Q,(0)}=I-D_U^Q\).

\begin{proposition}[Derived hybrid-path stability]
\label{prop:derived-hybrid-path-stability}
Let
\begin{align*}
a_*&=\max_{U\subseteq\mathcal U}
\|(I-B^U)^{-1}\|_{\mathrm{op}},\\
\delta_B&=\sum_{r=0}^1
\|\widehat B^{(r)}-B^{(r)}\|_{\mathrm{op}}.
\end{align*}
For \(\Theta_s=\Theta+s(\widehat\Theta-\Theta)\), let \(B_s^U\) be its
coalition-\(U\) hybrid.  If \(a_*\delta_B<1\), then
\begin{align}
\max_{U,s}\|(I-B_s^U)^{-1}\|_{\mathrm{op}}
&\le \frac{a_*}{1-a_*\delta_B}.
\label{eq:derived-hybrid-resolvent}
\end{align}
Consequently, the right side is at most \(2a_*\) with probability tending
to one whenever
\(\|\widehat B^{(r)}-B^{(r)}\|_{\mathrm F}=O_{\mathrm P}(\rho_r)\) and
\(a_*(\rho_0+\rho_1)=o(1)\).
\end{proposition}

\begin{proof}
Every \(B^U\) is strictly upper triangular, so \(a_*<\infty\).
Put \(E_B^{(r)}=\widehat B^{(r)}-B^{(r)}\).  By the definition of \(D_U^F\),
\begin{align*}
E_B^U:=\widehat B^U-B^U
&=E_B^{(0)}(I-D_U^F)+E_B^{(1)}D_U^F,\\
\|E_B^U\|_{\mathrm{op}}&\le\delta_B.
\end{align*}
Because \(\mathcal B\) contains only admissible ordered slopes, every
\(B_s^U\) is also strictly upper triangular.  Thus every path resolvent
exists without a small-error condition; \(a_*\delta_B<1\) is used only for
the following uniform operator-norm control.
Writing \(A^U=(I-B^U)^{-1}\), factor
\[
I-B_s^U=(I-B^U)(I-sA^UE_B^U).
\]
The Neumann bound gives
\(\|(I-sA^UE_B^U)^{-1}\|_{\mathrm{op}}
\le(1-a_*\delta_B)^{-1}\), proving
\eqref{eq:derived-hybrid-resolvent}.  Finally,
\(\delta_B\le\sum_r\|E_B^{(r)}\|_{\mathrm F}
=O_{\mathrm P}(\rho_0+\rho_1)\), so \(a_*\delta_B=o_{\mathrm P}(1)\).
\end{proof}

\begin{lemma}[Endpoint-gradient representation]
\label{lem:player-specific-gradient-representation}
Write \(\theta_s^U=(B_s^U,\mu_s^U,\omega_s^U)\) for the hybrid state
formed from \(\Theta_s\), and put
\[
w_U=\frac{|U|!(|\mathcal U|-|U|-1)!}{|\mathcal U|!}.
\]
At any hybrid state, with \(A=(I-B)^{-1}\), the score gradients are
\begin{align}
\nabla_\mu T
&=2AWA^\top\mu+Az,\notag\\
\nabla_\omega T
&=\diag(AWA^\top),\notag\\
\nabla_B T
&=A^\top\{2(\Diag(\omega)+\mu\mu^\top)AW
+\mu z^\top\}A^\top.
\label{eq:plugin-gradients}
\end{align}
For \(r\in\{0,1\}\) and \(\vartheta\in\{\mu,\omega\}\), the endpoint
gradients of the player-\(u\) attribution satisfy
\begin{align}
\nabla_{B^{(r)}}\Phi_u(\Theta_s)
&=\sum_{U\subseteq\mathcal U\setminus\{u\}}w_U
\{\nabla_BT(\theta_s^{U\cup\{u\}})D_{U\cup\{u\}}^{F,(r)}
\notag\\[-2pt]
&\hspace{29mm}
-\nabla_BT(\theta_s^U)D_U^{F,(r)}\},\notag\\
\nabla_{\vartheta^{(r)}}\Phi_u(\Theta_s)
&=\sum_{U\subseteq\mathcal U\setminus\{u\}}w_U
\{D_{U\cup\{u\}}^{Q,(r)}
\nabla_\vartheta T(\theta_s^{U\cup\{u\}})
\notag\\[-2pt]
&\hspace{29mm}
-D_U^{Q,(r)}\nabla_\vartheta T(\theta_s^U)\}.
\label{eq:player-specific-gradient-representation}
\end{align}
\end{lemma}

\begin{proof}
Let \(S=\Diag(\omega)+\mu\mu^\top\).  The score functional can be written as
\[
T(B,\mu,\omega)=\mathrm{tr}(SAWA^\top)+\mu^\top Az.
\]
For fixed \(B\), differentiation with respect to \(\mu\) and \(\omega\)
immediately gives the first two lines of~\eqref{eq:plugin-gradients}.
For a slope perturbation, \(dA=A(dB)A\).  Applying this identity to both
occurrences of \(A\), using symmetry of \(S\) and \(W\), and cycling the
trace gives
\[
d_BT
=\left\langle
A^\top\{2SAW+\mu z^\top\}A^\top,dB
\right\rangle_{\mathrm F},
\]
which proves the third line.

When only endpoint \(r\) is perturbed, the hybrid masks give
\begin{align*}
dB_s^U&=(dB^{(r)})D_U^{F,(r)},\\
d\mu_s^U&=D_U^{Q,(r)}d\mu^{(r)},&
d\omega_s^U&=D_U^{Q,(r)}d\omega^{(r)}.
\end{align*}
The Shapley value is the finite sum
\[
\Phi_u(\Theta_s)
=\sum_{U\subseteq\mathcal U\setminus\{u\}}w_U
\{T(\theta_s^{U\cup\{u\}})-T(\theta_s^U)\}.
\]
Differentiating it term by term and applying the three mask identities proves
\eqref{eq:player-specific-gradient-representation}.  In particular, the
before/after endpoint gradients are subtracted before any norm is taken.
\end{proof}

Equation~\eqref{eq:player-specific-gradient-representation} is the exact
player-specific formula: endpoint directions common to the two marginal
values cancel.  The exact sensitivities can therefore be evaluated by
differentiating the moment-and-quadrature program.

For deterministic rate notation, put
\[
q_{\mu,r}=\nu_r+(M_r+\nu_r)\rho_r,\qquad
q_{\omega,r}=\nu_r+\rho_r^2,
\]
and define
\begin{align*}
m_*&=1+\sum_{r=0}^1
\{\|\mu^{(r)}\|_2+\sqrt d\,q_{\mu,r}\},\\
\omega_*&=1+\max_r
\{\|\omega^{(r)}\|_\infty+q_{\omega,r}\},\\
H_B&=16a_*^3(\sqrt d\,\omega_*+m_*^2)W_*+4a_*^2m_*z_*,\\
H_\mu&=\sqrt d\{8a_*^2W_*m_*+2a_*z_*\},\qquad
H_\omega=4da_*^2W_*.
\end{align*}

\begin{lemma}[Uniform endpoint-gradient rate]
\label{lem:uniform-player-sensitivity-rates}
Under the conditions of
Corollary~\ref{cor:regime-residual-moment-rates} in both regimes, assume
Assumption~\ref{ass:residual-moments},
\(\nu_r=o(1)\), and \(a_*(\rho_0+\rho_1)=o(1)\).  Then
\begin{align}
\max_{u,r}L_{u,\vartheta}^{(r)}
&=O_{\mathrm P}(H_\vartheta),
&&\vartheta\in\{B,\mu,\omega\}.
\label{eq:deterministic-global-sensitivity-rates}
\end{align}
\end{lemma}

\begin{proof}
Put
\(\alpha_{\mathrm{path}}=\sup_{U,s}\|(I-B_s^U)^{-1}\|_{\mathrm{op}}\),
\(m_{\mathrm{path}}=\sup_{U,s}\|\mu_s^U\|_2\), and
\(\omega_{\mathrm{path}}=\sup_{U,s}\|\omega_s^U\|_\infty\).
The coordinate masks and endpoint interpolation give
\begin{align*}
m_{\mathrm{path}}&\le\sum_{r=0}^1
\{\|\mu^{(r)}\|_2+\sqrt d\,\eta_\mu^{(r)}\},\\
\omega_{\mathrm{path}}&\le\max_{r=0,1}
\{\|\omega^{(r)}\|_\infty+\eta_\omega^{(r)}\}.
\end{align*}
Corollary~\ref{cor:regime-residual-moment-rates} gives
\(m_{\mathrm{path}}=O_{\mathrm P}(m_*)\) and
\(\omega_{\mathrm{path}}=O_{\mathrm P}(\omega_*)\), while
Proposition~\ref{prop:derived-hybrid-path-stability} gives
\(\alpha_{\mathrm{path}}\le2a_*\) with probability tending to one.
Moreover,
\begin{align*}
\|\Diag(\omega)+\mu\mu^\top\|_{\mathrm F}
&\le\sqrt d\|\omega\|_\infty+\|\mu\|_2^2,\\
\|v\|_1&\le\sqrt d\|v\|_2,\\
\|\diag(C)\|_1&\le d\|C\|_{\mathrm{op}}.
\end{align*}
Applying these inequalities to~\eqref{eq:plugin-gradients} bounds the three
score-gradient norms by the corresponding
\(H_B,H_\mu,H_\omega\) terms, up to universal constants.
The masks and \(\Pi_{\mathcal B}\) are contractive, the Shapley weights sum
to one, and each marginal contribution contains two score gradients.
The triangle inequality in
\eqref{eq:player-specific-gradient-representation} proves the result
uniformly over the prespecified player universe.  This rate envelope discards the
before/after cancellation retained by the exact \(L\)'s.
\end{proof}

\begin{proof}[Proof of Theorem~\ref{thm:population-plugin-rate}]
Put \(E_\vartheta^{(r)}
=\widehat\vartheta^{(r)}-\vartheta^{(r)}\) for
\(\vartheta\in\{B,\mu,\omega\}\).
Because \(\mathcal B\) contains only admissible ordered slopes, every hybrid
on \(\Theta_s\) remains strictly upper triangular, so the endpoint functional
is differentiable along the entire path.
Lemma~\ref{lem:player-specific-gradient-representation} gives its endpoint
gradients.
The fundamental theorem of calculus therefore yields
\begin{align*}
\widehat\Phi_u-\Phi_u
=\int_0^1\sum_{r=0}^1\Bigl[
&\langle\nabla_{B^{(r)}}\Phi_u(\Theta_s),
E_B^{(r)}\rangle_{\mathrm F}\\
&+\nabla_{\mu^{(r)}}\Phi_u(\Theta_s)^\top E_\mu^{(r)}\\
&+\nabla_{\omega^{(r)}}\Phi_u(\Theta_s)^\top E_\omega^{(r)}
\Bigr]\,ds .
\end{align*}
Since \(E_B^{(r)}\in\mathcal B\), its inner product is unchanged after
projecting the gradient by \(\Pi_{\mathcal B}\).  Frobenius
Cauchy--Schwarz and \(\ell_1/\ell_\infty\) duality then prove
\eqref{eq:player-specific-deterministic-bound}.

For the stochastic part,
Proposition~\ref{prop:derived-hybrid-path-stability} gives the
\(2a_*\) resolvent bound, so estimated-path stability is derived rather than
assumed.
Corollary~\ref{cor:regime-residual-moment-rates}, applied conditionally on
each mechanism fit, gives the \(q_{\mu,r}\) and \(q_{\omega,r}\) endpoint
rates.  Lemma~\ref{lem:uniform-player-sensitivity-rates} gives the
uniform \(H_\vartheta\) sensitivity rates.  Combining these results with
\(\eta_B^{(r)}=O_{\mathrm P}(\rho_r)\) in the deterministic inequality and maximizing
over the prespecified players gives
\(\|\widehat\Phi-\Phi\|_\infty=O_{\mathrm P}(R_d)\), with \(R_d\) defined
in~\eqref{eq:player-specific-plugin-rate}.

For fold-specific games targeting the same population endpoints, applying
the deterministic inequality to each fold and then the triangle inequality
bounds the averaged attribution error by the average of the foldwise bounds.
For a fixed number of folds, this preserves the displayed stochastic order.
Each training fold must satisfy its score, RE/RSC, radius, and tolerance
conditions at its own training-sample size; the folds themselves need not be
independent.
\end{proof}

\begin{proof}[Proof of Corollary~\ref{cor:ranking-recovery}]
Theorem~\ref{thm:population-plugin-rate} gives
\(\max_u|\widehat\Phi_u-\Phi_u|=O_{\mathrm P}(R_d)\).
For \(u\in S_k\) and \(v\notin S_k\),
\[
|\widehat\Phi_u|-|\widehat\Phi_v|
\ge |\Phi_u|-|\Phi_v|
-|\widehat\Phi_u-\Phi_u|-|\widehat\Phi_v-\Phi_v|.
\]
Since every population boundary difference is at least \(g_k\), the
condition \(R_d=o(g_k)\) makes the last two terms \(o_{\mathrm P}(g_k)\), uniformly
over these pairs.
Every estimated top-\(k\) magnitude therefore exceeds every remaining
magnitude with probability tending to one.  The same inequality against
zero proves the stated sign claim.  This argument compares the same
prespecified player universe and does not justify data-dependent screening.
\end{proof}

\section{Evaluation Details and Reproducibility}
\label{sec:app_experiments}

This appendix documents the constructions, protocols, and diagnostics behind
the main-paper example and experiments.  Sections~D.2--D.5 provide controlled
evidence for resolution, computation, scarce-incident estimation, and the
main modeling boundaries; Sections~D.6--D.7 examine measured interventions.
Unless stated otherwise, S$^4$-RCA denotes the flat function/law-resolved
report.  Synthetic results are means over independently generated problems,
with standard errors across problems.

\subsection{Common protocol and metrics}
\label{app:common-protocol}
For attribution magnitudes \(a\) and a known changed-player set \(G\), we use
\(\mathrm{Recall@}|G|=|G\cap\mathrm{Top}_{|G|}(a)|/|G|\).  When the target is
an oracle ranking rather than a known intervention, we instead report
Top-\(k\) agreement,
\(|\mathrm{Top}_k(\widehat\Phi)\cap\mathrm{Top}_k(\Phi)|/k\), and do not call
it changed-support recall.  In Sec.~\ref{app:estimation-experiments}, \(F\)
and \(Q\) Recall@5 rank each family separately, whereas \(F+Q\) Recall@10
ranks all \(2d\) typed players jointly.  Main Figure~2 evaluates the
function-only \(\lvert\Delta B\rvert\) proxy only by \(F\) Recall.  For the broad
supplemental comparison, every method that returns one node ranking assigns
both labels to each of its top five nodes.  This preserves the ten-label
prediction budget but is only a common localization diagnostic: lifted family
recalls do not imply function/law identification.
For measured interventions, Hit@\(k\) uses the recorded intervention target
as a localization label.  Worst-over-split ranks quantify partition
sensitivity within one recorded system, not independent-system uncertainty.

\subsection{Function/law resolution and action ambiguity}
\label{app:fq-resolution-example}

\paragraph{Purpose and source SCMs.}
The construction isolates information lost by a paired-node report: whether
the candidate response should restore a changed exogenous law or block an
incoming mechanism.  We generate 100 zero-mean, ordered
linear-Gaussian source SCMs with \(d=20\).  Six upstream nodes form an ordered
DAG with edge probability 0.25, independently signed coefficient magnitudes
\(\mathrm{Unif}(0.10,0.35)\), and one to five randomly chosen target parents.
The target and a distractor each feed a downstream chain of length zero to
three with coefficient magnitudes \(\mathrm{Unif}(0.25,0.75)\).  For a branch
of size \(m\), its quadratic-score block is
\(R^\top R/m+\operatorname{Diag}(h)\), with standard-normal
\(R\in\mathbb R^{m\times m}\) and
\(h_j\sim\mathrm{Unif}(0.25,0.75)\).  Here \(R^\top R/m\) introduces
positive-semidefinite cross-coordinate score interactions, while \(h\) gives
every branch coordinate a positive diagonal weight.  The block is
trace-normalized to \(m\),
weighted by \(\mathrm{Unif}(0.70,1.30)\), and has no cross-branch terms.
Upstream residual variances follow \(\mathrm{Unif}(0.70,1.30)\), and other
background variances follow \(\mathrm{Unif}(0.04,0.12)\).

\paragraph{Matched incident endpoints.}
Each source SCM is paired with two incident endpoints whose target
attributions are
\((\Phi_F,\Phi_Q)=(-\beta,+\alpha)\) in the Q-actionable case and
\((+\alpha,-\beta)\) in its mirror.  A distractor has
\(\Phi_Q=\delta\), with
\(\alpha>\beta>\delta>\alpha-\beta>0\).  Thus the target node has paired
attribution \(\alpha-\beta\), below the distractor in both endpoints, although
the appropriate target action differs.

To realize these identities, for each source SCM draw
\(\alpha\sim\mathrm{Unif}(1.5,2.5)\),
\(\beta/\alpha\sim\mathrm{Unif}(0.65,0.78)\),
\(\gamma/\alpha\sim\mathrm{Unif}(0.15,0.30)\), and
\(\delta\sim\mathrm{Unif}(\alpha-\beta+0.10\alpha,
\alpha-\beta+0.20\alpha)\).  A standard-normal target coefficient direction
is scaled to propagated score mass \(\beta+\gamma\).  Let \(s_t\) and \(s_r\)
be the exact score sensitivities to the target and distractor residual
variances.  The Q-actionable endpoint scales the target column by
\(\sqrt{\gamma/(\beta+\gamma)}\) and
adds \(\alpha/s_t\) and \(\delta/s_r\) to their respective variances.  The
mirror scales the target column by
\(\sqrt{(\beta+\gamma+\alpha)/(\beta+\gamma)}\) and subtracts
\(\beta/s_t\) from target variance, with the same distractor shift.  The
source target and distractor law-score masses are respectively
\(\beta+\mathrm{Unif}(0.40\alpha,0.80\alpha)\) and
\(\mathrm{Unif}(0.30\alpha,0.60\alpha)\).  No repeat is filtered by its
attribution or repair outcome.  Pair members have identical source
SCMs, scores, paired reports, and incident score contrasts.  The 100 source
problems therefore yield 100 matched pairs, or 200 incident endpoints.

\paragraph{Candidate actions and evaluation.}
In the Q-actionable task, a rule restores either the target or distractor law
to its source value.  The resolved rule ranks the corresponding \(Q\)
attributions, whereas the paired rule acts on its top-ranked node.  We report
the target selection rate, the selected score reduction
\(T(M^{(1)})-T(M^{(1),a})\), the post-repair shift
\(T(M^{(1),a})-T(M^{(0)})\), and alarm clearance when the latter is
nonpositive.  The mirror compares target-law restoration with blocking all
incoming target edges.  Since the paired reports are identical across the two
endpoints, any paired-only action rule has balanced accuracy at most 0.5.

\begin{table}[!t]
\centering
\small
\setlength{\tabcolsep}{4pt}
\begin{tabular}{@{}lrr@{}}
\toprule
Metric & Function/law-resolved & Paired node \\
\midrule
Target-Q selected (\%) & 100.0 & 0.0 \\
Selected score reduction & 1.934$\pm$0.029 & 0.838$\pm$0.016 \\
Post-repair score shift & -0.543$\pm$0.018 & 0.552$\pm$0.011 \\
Alarm cleared (\%) & 100.0 & 0.0 \\
\bottomrule
\end{tabular}

\caption{Q-actionable repair localization over 100 independent source
problems (mean $\pm$ standard error).  Score reduction is
\(T(M^{(1)})-T(M^{(1),a})\); post-repair shift is
\(T(M^{(1),a})-T(M^{(0)})\), so a nonpositive value clears the alarm.}
\label{tab:fq-actionability}
\end{table}

Table~\ref{tab:fq-actionability} shows that the resolved rule
always selects the target law and clears the alarm, whereas the paired rule
always selects the distractor and does not.  Across both matched endpoints,
the resolved action rule has balanced accuracy 1.0, whereas constant-Q,
constant-F, and paired-only rules attain or
are bounded by 0.5.  This oracle construction demonstrates repair ambiguity
under aggregation, not physical function/law identification or generic repair
optimization.

\subsection{Exact computation: scalability and numerical agreement}
\label{app:computation-experiments}

\paragraph{Question and protocol.}
We ask whether the specialized algorithm removes the combinatorial Shapley
bottleneck while preserving the same attribution.  All methods target the
same generated game; generic solvers receive its common coalition-value
callable.  The comparison therefore isolates Shapley aggregation rather than
endpoint estimation.  Stochastic solvers are included only to show the
accuracy--runtime tradeoff.
For each repetition and size, we generate a sparse ordered DAG with
independent zero-mean disturbances, a diagonal quadratic score, and all
\(K=2d\) function/law players.  The primary grid is
\(K\in\{10,20,40,100,200,400,1000\}\), and exact solvers also use
\(K\in\{4,6,8,10,14,20\}\).  We report mean \(\pm\) standard error over three
independently generated problems for every non-timeout cell.  Each method has
a 1000-second cap; after its first timeout, that cell and larger sizes are not
rerun.  DoWhy 0.14 uses its public defaults: Exact, ExactFast,
25-permutation Permutation, 5\% EarlyStopping, and 5000-sample SubsetSampling
\citep{blobaum2024dowhygcm}.
Relative error is
\(\|\widehat\Phi-\Phi^{\mathrm{S4}}\|_2/\|\Phi^{\mathrm{S4}}\|_2\).
Runs use an Intel Core i7-1068NG7 CPU (8 logical cores, 16\,GiB RAM).
Algorithm times exclude problem construction and process startup; DoWhy's
default \texttt{n\_jobs=-1} may use all logical cores, whereas S$^4$-RCA is
single-process.

\paragraph{Results.}
Main Figure~\ref{fig:main-scalability} shows that S$^4$-RCA is over
\(100\times\) faster than every exact-mode comparator at the largest size
attempted by both methods, and it is the only solver to complete \(K=1000\)
within the cap.  Full subset enumeration and DoWhy Exact numerically agree
with S$^4$-RCA, while the stochastic modes exhibit the expected
accuracy--runtime tradeoff.  DoWhy ExactFast fails the agreement check at
\(K=20\), so its nominal mode should not be interpreted as verified exactness
for this game.

\subsection{Scarce-incident attribution recovery}
\label{app:estimation-experiments}

\paragraph{Question and synthetic design.}
This benchmark asks whether reference anchoring recovers the leading
function and law changes from scarce incident data.  We generate 30 ordered
Gaussian \(d=100\) SCM pairs and one 100-row incident pool per pair;
\(n_1\in\{20,40,60,80,100\}\) uses nested prefixes of that pool.  The order
\(0{:}99\) contains two disconnected 50-node
blocks; within-block edges are Bernoulli\((0.10)\), nonzero coefficients are
signed \(\mathrm{Unif}(0.20,0.50)\), and
\(\omega_j^{(0)}=\mathrm{Unif}(0.8,1.2)^2\).
Replicate parity determines which block is monitored.  Each block contains
five function targets and five disjoint law targets; shifts in the unmonitored
block are score-null matched decoys.  A function target rewires
\(r_j=\min\{p_j,a_j,\max(1,\lceil0.2p_j\rceil)\}\) predecessor edges, where
\(p_j\) is its number of existing within-block predecessors and \(a_j\) is the
number of available absent predecessors.  We set \(r_j\) existing
coefficients to zero and add \(r_j\) absent edges with independently signed
\(\mathrm{Unif}(0.20,0.50)\) coefficients.  For each scored \(Q\) target, its
initial variance increment matches the standalone
score effect of one scored function rewiring.  A common multiplier then
equalizes the exact full-game total absolute function and law attribution
mass; the five resulting variance multipliers are copied to the decoy block.
Across the 30 SCMs, the multipliers range from 1.08 to 2.49, and every oracle
\(F+Q\) top ten contains all ten scored targets.  Residual means are zero and
disturbances are independent Gaussian.  Family-specific Recall@5 treats only
the five score-relevant targets as positives.

\paragraph{Reference monitor and sample splitting.}
The reference size is \(n_0=10{,}000\).  A fixed split uses 8,000 rows to fit
the ordered reference mechanism and monitor, and 2,000 disjoint rows to
estimate \(\mu^{(0)},\omega^{(0)}\).  Writing \(b\) for the monitored block,
the score uses the full 50-dimensional block:
\[
T^2(x_b)=(x_b-\bar x_b)^\top
\left\{\widehat\Sigma_b+10^{-3}\operatorname{tr}(\widehat\Sigma_b)I/50\right\}^{+}
(x_b-\bar x_b),
\]
where \(\bar x_b,\widehat\Sigma_b\) use only the 8,000 fit rows; the additive
constant is omitted.

For each incident sample size, five shuffled outer folds use 80\% of the rows
to fit \(B^{(1)}_\ell\) and the held-out 20\% only to estimate
\(\mu^{(1)}_\ell,\omega^{(1)}_\ell\).  We average signed Shapley vectors over
folds before ranking.  Reference Lasso, Delta-Lasso, and incident-only fits
select nodewise penalties by ten-fold CV using only their mechanism-training
rows.

\paragraph{Comparators.}
Resolved S$^4$-RCA uses the reference-anchored Delta-Lasso endpoint;
incident-only Shapley refits the same resolved game directly from incident
data.  Paired S$^4$-RCA replaces \((F_j,Q_j)\) jointly and therefore returns a
node ranking rather than a change type.  The function-only
\(\lvert\Delta B\rvert\) proxy ranks
\(j\) by
\(\|\widehat B^{(1)}_{\cdot j}-\widehat B^{(0)}_{\cdot j}\|_2\) and is
evaluated natively by \(F\) Recall@5.  Observed z/variance sums the standardized
absolute mean change and absolute log-variance ratio of each measured
coordinate.  Observed PCA standardizes all 100 nodes, retains the smallest
reference eigenspace explaining 95\% of variance, and ranks node \(j\) by the
incident mean of \(|Z_{ij}(Z_iW)_j|\), where \(W\) is the retained-subspace
precision.  Residual local KL is a Gaussian marginal-residual proxy for the
local conditional divergence of \citet{budhathoki2021why}: it applies the
fitted reference mechanism to both regimes and ranks
\(D_{\rm KL}\{N(\widehat\mu_j^{(1)},\widehat\omega_j^{(1)})\|N(\widehat\mu_j^{(0)},
\widehat\omega_j^{(0)})\}\).

The broader supplemental benchmark adds methods with adjacent but distinct
outputs.  Fixed-\(F\) averages signed event Shapley values under the fitted
reference function; it is our quadratic expected-score counterpart of
fixed-SCM outlier attribution, not its native information-theoretic tail game
\citep{budhathoki2022causal}.  Linear-Gaussian CCA jointly switches fitted node
conditionals using the incident-only endpoint, while MRCCA uses DoWhy 0.14's
multiply-robust score estimator with five folds, five Shapley permutations, a
2,000-row source cap, and propensity clipping at 0.02
\citep{budhathoki2021why,quintasmartinez2024multiply}.  We convert DCI's
estimated difference-DAG edge matrix \(\widehat\Delta\) to the node score
\(j\mapsto\|\widehat\Delta_{\cdot j}\|_2\); score ties are broken
lexicographically by feature name \citep{wang2018direct}.  We rank nodes by
iSCAN's returned variance-ratio shift statistic rather than thresholding its
shifted-node set \citep{chen2023iscan}.  Our feature-shift SHAP baseline adapts
SHAP by treating observed-coordinate distribution switches as players, without
causal propagation \citep{lundberg2017unified}.  FANS and do-Shapley
remain conceptual comparisons in Table~\ref{tab:close-work-map}: the former
does not return a scalar score attribution, and the latter targets variable
interventions within one fixed SCM.

\paragraph{Protocol checks.}
Although the synthetic endpoints change only \(B\) and \(\omega\), fitted games estimate all
residual means and variances from held-out data; only oracle diagnostics
receive true endpoints.
We prune only modules outside the union-endpoint score ancestors or identical
across endpoints at tolerance \(10^{-10}\), then restore their zero entries.
This is exact null-player pruning, not target-informed screening.
Every main-figure cell has 30 completed SCMs.  For the extended comparison,
DCI uses at most 500 rows per regime, conditioning sets of size at most two,
and a 300-second per-cell cap; it has no \(n_1=100\) point and its smaller-size
means use only successful cells.  Other extended methods complete all 150
SCM--sample-size cells.

\begin{figure*}[!t]
  \centering
  \includegraphics[width=\textwidth]{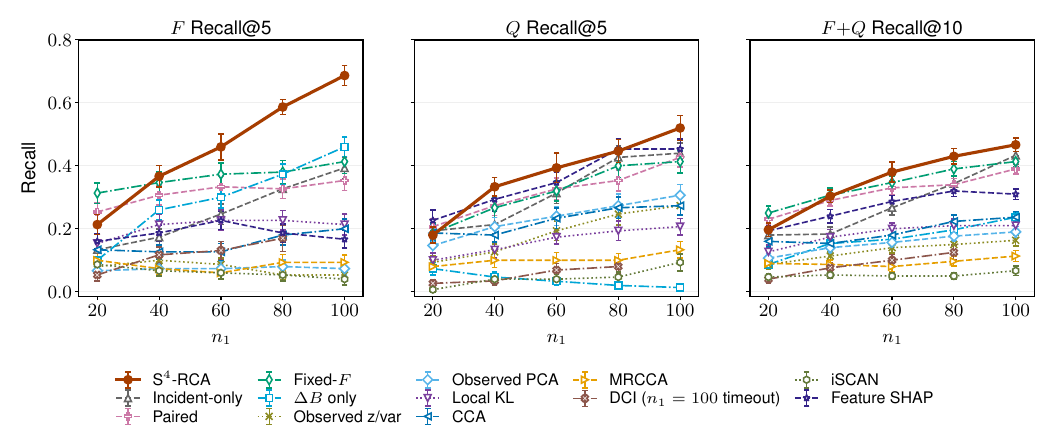}
  \caption{Broad comparison on the same 30 scarce-incident SCMs as Main
  Figure~2 (mean $\pm$ SE).  The main figure emphasizes transfer, resolution,
  and propagation ablations; this figure adds fixed-SCM event attribution,
  CCA/MRCCA, DCI, iSCAN, and feature-shift SHAP.  Single-ranking outputs are
  copied to both \(F\) and \(Q\) labels only for a common localization
  diagnostic and do not thereby identify change type.  Fixed-\(F\) is the
  expected-score event mean, not the native information-theoretic tail game.
  DCI has no \(n_1=100\) result because every such cell reached the 300-second
  cap.}
  \label{fig:extended-component-benchmark}
\end{figure*}

\paragraph{Results and interpretation.}
Main Figure~2 isolates the proposed method's components: reference anchoring
and score-aware propagation produce the clearest gains in function recovery,
while the paired ablation shows that joint localization alone does not retain
change type.  Supplement Figure~\ref{fig:extended-component-benchmark} places this result
among broader alternatives.  Fixed-\(F\) and feature-shift SHAP can be
competitive for law or untyped localization in some regimes, but neither
returns the resolved population report; CCA/MRCCA and the score-agnostic shift
detectors likewise answer coarser questions.  The comparison therefore
supports S$^4$-RCA when both score propagation and function/law resolution are
needed, rather than a universal claim across different estimands.

Table~\ref{tab:d100-crossfit-diagnostics} adds endpoint and full-vector error
diagnostics.  Recall improves with incident size even though full-vector error,
especially for the law family, remains substantial.  The evidence concerns
leading effects, not uniform magnitude recovery or calibrated rank
uncertainty.

\begin{table}[!t]
\centering
\begingroup
\small
\setlength{\tabcolsep}{2.4pt}
\begin{tabular*}{\columnwidth}{@{\extracolsep{\fill}}rccc@{}}
\toprule
$n_1$ & $F$ R@5 & $Q$ R@5 & $F{+}Q$ R@10 \\
\midrule
20 & 0.21\,(0.03) & 0.18\,(0.03) & 0.20\,(0.02) \\
40 & 0.37\,(0.03) & 0.33\,(0.03) & 0.30\,(0.02) \\
60 & 0.46\,(0.04) & 0.39\,(0.05) & 0.38\,(0.03) \\
80 & 0.59\,(0.02) & 0.45\,(0.04) & 0.43\,(0.02) \\
100 & 0.69\,(0.03) & 0.52\,(0.04) & 0.47\,(0.02) \\
\bottomrule
\end{tabular*}
\par\smallskip
\begin{tabular*}{\columnwidth}{@{\extracolsep{\fill}}rcccc@{}}
\toprule
$n_1$ & $F$ rel. $\ell_1$ & $Q$ rel. $\ell_1$ & $\|\widehat B-B\|_{\rm op}$ & $\|\widehat\omega-\omega\|_\infty$ \\
\midrule
20 & 6.84\,(0.30) & 12.80\,(0.44) & 1.90\,(0.06) & 4.27\,(0.18) \\
40 & 3.76\,(0.22) & 8.13\,(0.30) & 1.47\,(0.03) & 2.87\,(0.09) \\
60 & 2.18\,(0.16) & 5.86\,(0.28) & 1.15\,(0.03) & 1.94\,(0.08) \\
80 & 1.47\,(0.09) & 4.79\,(0.19) & 0.96\,(0.02) & 1.54\,(0.04) \\
100 & 1.12\,(0.06) & 4.13\,(0.17) & 0.76\,(0.02) & 1.27\,(0.03) \\
\bottomrule
\end{tabular*}
\endgroup

\caption{Calibrated \(d=100\) cross-fitted diagnostics over 30 SCMs; entries
are mean (SE). \(F\) and \(Q\) R@5 are family-specific changed-support
recalls, and \(F+Q\) R@10 ranks the 200 typed players jointly. Relative \(\ell_1\) errors
compare the family attribution vectors with their exact oracles. Endpoint
errors first average the five folds within each SCM.}
\label{tab:d100-crossfit-diagnostics}
\end{table}

\FloatBarrier
\subsection{Core-assumption stress tests}
\label{app:robustness-experiments}

The guarantees assume a supplied common order, linear additive mechanisms,
and a quadratic score.  We vary these three conditions separately over ten
independently generated \(d=5\) SCMs.  Each diagnostic has two layers:
\emph{Oracle} isolates the population attribution distortion caused by the
tested departure, whereas \emph{E2E} also estimates both SCM endpoints from
samples.  Relative \(\ell_1\) error compares the resulting ten-player vector
with the corresponding exact target, and R@\(k\) is changed-player recall.

All E2E runs use an 80/20 reference fit/moment split, reference Lasso and
Delta-Lasso with five-fold internal CV, five-fold incident cross-fitting with
held-out residual moments, and signed fold averaging.  We set \(n_1=500\) and,
for the score and mechanism tests, \(n_0=1000\); the order test varies
\(n_0\).  The supplied score remains fixed, so E2E adds endpoint-estimation
error rather than monitor-training error.  These diagnostics examine
misspecification, while Sec.~\ref{app:estimation-experiments} supplies the
scarce-incident evidence.  Table~\ref{tab:d5-assumption-stress} reports mean
(SE) over SCMs.

\paragraph{Reference-only order estimation.}
We generate unique-order linear SCMs with random variable relabeling. The
chain coefficients have independently random signs and magnitudes
\(\mathrm{Unif}(0.55,0.80)\); each other predecessor edge appears with
probability 0.45 and signed magnitude \(\mathrm{Unif}(0.15,0.35)\).
Independent centered Laplace disturbances have variances
\(\mathrm{Unif}(0.7,1.3)\), and the diagonal score weights are
\(\mathrm{Unif}(0.6,1.4)/d\). The incident endpoint multiplies the last chain
coefficient by 1.40 and the second-position disturbance variance by 1.60.
DirectLiNGAM from \texttt{lingam} 1.12.1 \citep{shimizu2011directlingam}
receives the 80\% fit portion of nested total reference samples with
\(n_0\in\{100,200,300,400\}\); the same fit portion estimates the reference
mechanisms in E2E. Oracle
reparameterizes the population source and target covariances under the learned
order, excluding endpoint sampling error.  Table~\ref{tab:d5-assumption-stress}
shows that improved order recovery removes the Oracle distortion, whereas E2E
error remains governed largely by endpoint estimation.  This is a favorable
linear non-Gaussian identification check, not generic order-recovery evidence.

\paragraph{Nonlinear score approximation.}
Each zero-mean ordered linear-Gaussian SCM has positive chain coefficients
\(B_{j-1,j}\sim\mathrm{Unif}(0.20,0.38)\); every other predecessor edge appears
with probability 0.35 and coefficient \(\mathrm{Unif}(0.08,0.20)\).
The independent disturbances have variances \(\mathrm{Unif}(0.75,1.25)\).
The incident endpoint scales incoming columns \(d-1\) and \(d\) of \(B\) by
1.30 and 0.72 and disturbance variances 2 and 3 by 1.50 and 1.35. In
reference-centered coordinates, define, for \(j\in[d]\),
\[
s_\eta(x)=x^\top W x+\eta\sum_j v_jx_j^4.
\]
Independently draw \(\widetilde w_j,\widetilde v_j\sim
\mathrm{Unif}(0.7,1.3)\), and set
\(W=\operatorname{Diag}(\widetilde w)/d\) and
\(v_j=\widetilde v_j/d>0\).  Thus both quadratic and quartic weights are
positive and have order-one total scale.  For reference covariance \(\Sigma_0\),
put \(q_0=\operatorname{tr}(W\Sigma_0)\) and
\(f_0=3\sum_{j=1}^d v_j\Sigma_{0,jj}^2\), and set \(\eta=rq_0/f_0\) for
\(r\in\{0.2,0.4,0.6\}\). Thus the quartic reference expectation is exactly
\(r\) times the quadratic one, and the degree-two Taylor approximation at the
zero reference center is \(x^\top W x\). Gaussian second and fourth moments
give all \(2^{2d}\) nonlinear-score coalition values exactly.  The table shows
that increasing quartic mass distorts attribution magnitude while leaving the
tested leading-player recall stable.  This smooth stress does not cover
arbitrary threshold or tail scores.

\paragraph{Nonlinear mechanism stress.}
We next use
\[
x_j=\sum_{i<j}B_{ij}x_i+\sum_{i<j}G_{ij}\tanh(x_i)+e_j
\]
on a chain-plus-extra-edge graph: chain \(B\) coefficients follow
\(\mathrm{Unif}(0.22,0.38)\), chain unit-\(G\) coefficients follow
\(\mathrm{Unif}(0.25,0.45)\), and each other predecessor is included with
probability 0.35 with coefficients
\(B_{ij}\sim\mathrm{Unif}(-0.16,0.16)\) and
\(G_{ij}^{\rm unit}\sim\mathrm{Unif}(-0.25,0.25)\). At strength
\(c\in\{0.3,0.6,0.9\}\), \(G^{(0)}=cG^{\rm unit}\); the incident endpoint
scales columns \(d-1,d\) of \(B\) by 1.30 and 0.72 and of \(G\) by 1.45 and
0.60. It also scales disturbance variances 2 and 3 by 1.50 and 1.35, from
independent Gaussian baselines \(\mathrm{Unif}(0.70,1.10)\); the diagonal
quadratic-score weights follow \(\mathrm{Unif}(0.7,1.3)/d\).
Tensor Gauss--Hermite integration of order nine numerically approximates every
nonlinear-hybrid coalition value, with order eleven as an audit. Population
weighted least squares on all predecessors in the true order supplies each
regime's linear reapproximation. These projections match the quadrature
endpoint means and covariances and hence the quadratic-score shift, but need
not preserve hybrid allocations.  The Oracle diagnostic exposes increasing
allocation error with nonlinearity, while E2E leading-player recovery remains
comparatively stable.  The retained ranking does not identify physical
nonlinear mechanisms.

\begin{table}[!t]
\centering
\begingroup
\small
\setlength{\tabcolsep}{4pt}
\begin{tabular*}{\columnwidth}{@{\extracolsep{\fill}}rrrr@{}}
\toprule
\multicolumn{4}{c}{Reference-only order estimation} \\
\cmidrule(lr){1-4}
$n_0$ & Oracle $\ell_1$ & E2E $\ell_1$ & E2E R@2 \\
\midrule
100 & 0.14 (0.09) & 2.11 (0.19) & 0.40 (0.12) \\
200 & 0.11 (0.11) & 1.63 (0.22) & 0.55 (0.09) \\
300 & 0.16 (0.16) & 1.67 (0.20) & 0.60 (0.07) \\
400 & 0.00 (0.00) & 1.12 (0.15) & 0.75 (0.08) \\
\bottomrule
\end{tabular*}

\par\smallskip
\begin{tabular*}{\columnwidth}{@{\extracolsep{\fill}}rrrr@{}}
\toprule
\multicolumn{4}{c}{Nonlinear-score approximation} \\
\cmidrule(lr){1-4}
Quartic ratio & Oracle $\ell_1$ & E2E $\ell_1$ & E2E R@4 \\
\midrule
20\% & 0.32 (0.01) & 0.60 (0.03) & 0.68 (0.04) \\
40\% & 0.49 (0.01) & 0.70 (0.02) & 0.68 (0.04) \\
60\% & 0.59 (0.01) & 0.75 (0.02) & 0.68 (0.04) \\
\bottomrule
\end{tabular*}

\par\smallskip
\begin{tabular*}{\columnwidth}{@{\extracolsep{\fill}}rrrr@{}}
\toprule
\multicolumn{4}{c}{Nonlinear-mechanism stress} \\
\cmidrule(lr){1-4}
Strength & Oracle $\ell_1$ & E2E $\ell_1$ & E2E R@4 \\
\midrule
0.3 & 0.03 (0.00) & 0.64 (0.05) & 0.57 (0.04) \\
0.6 & 0.07 (0.00) & 0.63 (0.05) & 0.65 (0.04) \\
0.9 & 0.12 (0.01) & 0.61 (0.05) & 0.72 (0.04) \\
\bottomrule
\end{tabular*}

\endgroup
\caption{Core-assumption diagnostics over ten \(d=5\) SCMs (mean (SE)). The
three blocks vary reference size, nonlinear-score quartic ratio, and
nonlinear-mechanism strength. Oracle \(\ell_1\) isolates population distortion
from the learned order or linear approximation; E2E \(\ell_1\) additionally
includes endpoint estimation and cross-fitting. R@\(k\) is changed-player
recall.}
\label{tab:d5-assumption-stress}
\end{table}

Zero-departure controls agree numerically, and the higher-order
Gauss--Hermite audit leaves the nonlinear-mechanism conclusion unchanged.
With ten SCMs, these tests expose qualitative boundaries rather than support a
general robustness claim.  Latent confounding, cycles, dependent disturbances,
and general nonlinear or nonadditive identification remain outside scope.

\FloatBarrier
\subsection{Causal Chamber: measured Light-Tunnel interventions}
\label{app:causal-chamber-core}

\paragraph{Question, data, and score.}
We ask whether a resolved report can separate the manipulated target-law
component of a fixed all-channel alarm from propagated fitted-response
components.  We use 16 Light-Tunnel set-point interventions whose manipulated
variable is among the 22 measured channels \citep{gamella2025chamber}.
Disjoint reference subsets fit a 12-component PCA \(T^2+\mathrm{SPE}\)
monitor (3,500 rows), calibrate its two terms (1,500), fit the ordered SCM
(3,500), and estimate residual moments (340); 12 is the smallest
reference-only dimension explaining 90\% of the variance.  The fixed score is
\(T^2/\operatorname{sd}_{\rm cal}(T^2)
+\mathrm{SPE}/\operatorname{sd}_{\rm cal}(\mathrm{SPE})\), with both scales
estimated on the disjoint calibration split.  It includes all nonconstant
measured channels, including setpoints, because the task is to allocate the
change in a deployed operational score rather than infer a hidden target.
A 300-row incident update split supplies five prespecified shuffled two-fold
partitions, and a disjoint 400-row split audits the monitor
(mean/median/minimum AUROC \(0.819/0.797/0.644\)).
The five repetitions perturb the partition of each recording rather than
provide independent physical experiments.
We restrict the official Light-Tunnel topological order to the 20 mapped
channels and append \texttt{v\_board} and \texttt{v\_reg}, which are absent
from the official graph, in their prespecified final positions.

\paragraph{Methods, metrics, and checks.}
S$^4$-RCA ranks the flat report by
\(a_j=|\overline\Phi_{F_j}|+|\overline\Phi_{Q_j}|\), averaging signed fold
values before taking absolute mass.  The Fixed-\(F\) control holds \(B\) at
its reference fit and updates residual moments on the same held-out rows; it
is a restricted endpoint, not a claim of \(Q\)-only physical truth.
Observed PCA instead ranks changes in mean per-channel score contribution on
the same 300-row incident update split relative to the disjoint 340-row
reference moment split; the independent 400-row incident split is used only
for monitor AUROC.
The manipulated channel is the localization label, and reported ranks are the
worst across the five partitions.  The primary S$^4$-RCA endpoint permits all
predecessors under the supplied order; official-edge and reference-fit
supports are sensitivity analyses.

\paragraph{Localization and resolved allocation.}
Main Figure~3 reports target ranks and the harder polarization allocations.
Across all 16 interventions, S$^4$-RCA ranks the target node and its \(Q_j\)
component first, invariant to the tested split and support variants.  In the
polarization cases, target-law mass is separated from predominantly downstream
\(F\) mass.  This is evidence for resolved allocation of an all-channel score,
not hidden-target localization.

\paragraph{Optical interpretation and boundary.}
The supplied topology places the \texttt{ir\_3} infrared sensor downstream of
both polarizer controls.  Their measured relative set-point angle moves from
near parallel to about \(\pm40^\circ\) and \(\pm79^\circ\), substantially
reducing the corresponding Malus factor
\(\cos^2(\theta_1-\theta_2)\).  This provides a physical rationale for
regime-specific linear projections to \texttt{ir\_3}, but does not establish
that the optical law itself changed.  The resolved allocation in Main
Figure~3 and the projection audit below are consistent with this
relinearization interpretation.

\paragraph{Projection audit.}
Table~\ref{tab:causal-chamber-optical-geometry} replaces the dense graph
display with the coefficients relevant to the optical interpretation.
The supplied topology is fixed, and coefficients are all-row OLS projections
in reference-standardized coordinates.  Mid-regime signs mirror across the
manipulated polarizer, while strong-regime \(R^2\) declines.  These patterns
support the relinearization interpretation but establish neither a changed
physical DAG nor a changed physical coefficient.

\begin{table}[!t]
\centering
\begingroup
\small
\setlength{\tabcolsep}{3.5pt}
\renewcommand{\arraystretch}{0.96}
\begin{tabular*}{\columnwidth}{@{\extracolsep{\fill}}lrrrr@{}}
\toprule
\multicolumn{5}{c}{Polarizer coefficients} \\
\cmidrule(lr){1-5}
 & \multicolumn{2}{c}{\(\to\) \texttt{ir\_3}} &
\multicolumn{2}{c}{\(\to\) \texttt{vis\_3}} \\
\cmidrule(lr){2-3}\cmidrule(l){4-5}
Regime & P1 & P2 & P1 & P2 \\
\midrule
Baseline & \(-0.05\) & \(+0.05\) & \(-0.05\) & \(+0.04\) \\
P1-mid   & \(-0.38\) & \(+0.35\) & \(-0.36\) & \(+0.34\) \\
P1-strong& \(-0.10\) & \(+0.10\) & \(-0.08\) & \(+0.09\) \\
P2-mid   & \(+0.37\) & \(-0.38\) & \(+0.35\) & \(-0.36\) \\
P2-strong& \(+0.17\) & \(-0.18\) & \(+0.16\) & \(-0.17\) \\
\bottomrule
\end{tabular*}
\par\smallskip
\begin{tabular*}{\columnwidth}{@{\extracolsep{\fill}}lrr@{}}
\toprule
\multicolumn{3}{c}{Projection \(R^2\)} \\
\cmidrule(lr){1-3}
Regime & \texttt{ir\_3} & \texttt{vis\_3} \\
\midrule
Baseline & 0.873 & 0.879 \\
P1-mid   & 0.861 & 0.863 \\
P1-strong& 0.438 & 0.457 \\
P2-mid   & 0.866 & 0.867 \\
P2-strong& 0.655 & 0.663 \\
\bottomrule
\end{tabular*}
\endgroup

\caption{Reference-standardized polarization projections on the fixed
official subgraph. P1/P2 are \texttt{pol\_1}/\texttt{pol\_2}; coefficients
and \(R^2\) use all rows and are descriptive rather than held-out evidence.}
\label{tab:causal-chamber-optical-geometry}
\end{table}

\subsection{External intervention stress tests: Wind Tunnel and Sachs}
\label{app:external-real-data}

\paragraph{Purpose, methods, and metric.}
We use Wind Tunnel to test whether the all-channel operational-score analysis
extends to a second physical system, and Sachs to examine the same pipeline
under substantially weaker model assumptions.  We compare S$^4$-RCA,
Fixed-\(F\), raw endpoint-shift magnitude, Observed PCA contribution, and
Residual local KL; an order-only S$^4$-RCA fit is a support sensitivity.  Raw
endpoint shift combines the fitted \(B\)-column change with standardized
residual-location and log-variance changes, without score propagation.  A
condition counts as Hit@\(k\) only when at least one documented target ranks
within the top \(k\) in all five split repetitions.

\begin{table}[!t]
\centering
\begingroup
\small
\setlength{\tabcolsep}{2.2pt}
\begin{tabular*}{\columnwidth}{@{\extracolsep{\fill}}lrrr@{}}
\toprule
Method & H@1 & H@3 & Med. \\
\midrule
\multicolumn{4}{@{}l}{\textbf{(a) Wind Tunnel}} \\
\addlinespace[1pt]
S$^4$-RCA (function/law) & 0.808 & \textbf{1.000} & \textbf{1.0} \\
Fixed-$F$ control & 0.769 & \textbf{1.000} & \textbf{1.0} \\
Delta shift & \textbf{1.000} & \textbf{1.000} & \textbf{1.0} \\
Observed PCA & 0.731 & 0.923 & \textbf{1.0} \\
Residual local KL & \textbf{1.000} & \textbf{1.000} & \textbf{1.0} \\
S$^4$-RCA (function/law; order-only) & 0.731 & 0.962 & \textbf{1.0} \\
\midrule
\multicolumn{4}{@{}l}{\textbf{(b) Sachs}} \\
\addlinespace[1pt]
S$^4$-RCA (function/law) & 0.143 & \textbf{0.429} & \textbf{6.0} \\
Fixed-$F$ control & 0.143 & 0.286 & 8.0 \\
Delta shift & 0.143 & 0.286 & 9.0 \\
Observed PCA & \textbf{0.286} & 0.286 & \textbf{6.0} \\
Residual local KL & 0.143 & 0.286 & 8.0 \\
S$^4$-RCA (function/law; order-only) & 0.143 & \textbf{0.429} & \textbf{6.0} \\
\bottomrule
\end{tabular*}
\endgroup

\caption{All-channel external interventions. H@\(k\) requires a target to
rank within the top \(k\) in all five splits; Med. is the median worst rank
across 26 Wind or seven Sachs conditions. Primary rows use official-edge
support, and order-only S$^4$-RCA is a sensitivity. Bold marks the
within-dataset best, including ties.}
\label{tab:external-real-data}
\end{table}

\paragraph{Wind Tunnel.}
We retain all 26 nonempty 10,000-row single-target regimes in
\texttt{wt\_walks\_v1}, covering 11 targets and 26 nonconstant channels after
excluding six constants \citep{gamella2025chamber}.
Disjoint reference subsets fit/calibrate an all-channel PCA
\(T^2+\mathrm{SPE}\) score, the ordered SCM, and residual moments. The five
reference roles use 3,000/1,500/3,000/1,000/1,500 rows for PCA,
score calibration, SCM fitting, residual moments, and monitor evaluation;
the selected monitor has six components.  Five shuffled two-fold splits use
300 update rows, and 1,500 disjoint rows audit each incident.  Mean/minimum
AUROC is \(0.974/0.761\).

Table~\ref{tab:external-real-data} shows that direct endpoint and residual-KL
detectors localize these single-target interventions at least as well as the
score-attribution methods.  S$^4$-RCA nevertheless places every target in the
top three and assigns little mass to \(F\), supporting stable, predominantly
\(Q\)-shift allocation rather than superiority over direct shift detectors.

\paragraph{Sachs.}
We log-transform 11 phosphoproteins and use condition 1 as reference; the
seven tests are AKT inhibitor, G06976, Psitectorigenin, U0126, LY294002, PMA,
and the dataset label \texttt{b2cAMP} \citep{sachs2005causal}.  Labels are documented measured targets,
except that measured Akt serves as a proxy for unmeasured PI3K in LY294002,
and, for U0126, either measured MEK, the direct intervention target, or
downstream ERK counts as a hit.
The 17-edge consensus DAG supplies support/order.  Disjoint reference subsets
of 200/150/250/126/127 cells serve PCA fitting, score calibration, SCM fitting,
residual-moment estimation, and monitor evaluation, respectively; the monitor
has eight components, and 400/200 incident cells are used for endpoint updates
and evaluation.
Sachs mean/minimum monitor AUROC is \(0.844/0.630\), and only four conditions
reach 0.75.  No method in Table~\ref{tab:external-real-data} localizes more
than three of the seven targets within the top three.  We retain all seven in
the common denominator; the weak monitor and localization make Sachs an
assumption-boundary stress test, not positive cross-domain validation.

\paragraph{Interpretation boundary.}
Light and Wind allocate a fixed all-channel score under a supplied ordered
linear reapproximation; endpoint shift, Residual local KL, and Observed PCA do
not share the resolved function/law estimand.  Sachs additionally involves
proxy targets, latent confounding, and nonlinear biology.  These studies do
not establish hidden-target localization, causal discovery, physical
mechanism recovery, universal superiority, or calibrated rank uncertainty.

\subsection{Reproducibility}
\label{app:repro}
\paragraph{Software and traceability.}
Experiments use Python 3.12.4 with NumPy 2.3.5, SciPy 1.15.3, scikit-learn
1.8.0, pandas 2.3.3, and Matplotlib 3.10.8; the solver audit additionally uses
DoWhy 0.14, the extended benchmark uses \texttt{causaldag} 0.1a163 and
\texttt{iscan-dag} 1.0.0, and the order diagnostic uses \texttt{lingam}
1.12.1. Internal
workflows retain raw and aggregated outputs and run manifests linking
commands, settings, seeds, and paper artifacts. The
principal synthetic seeds are 270730 for the action-ambiguity construction,
270713 for computation, 20270629 for scarce-incident recovery, and 20260728
for the assumption diagnostics.  Light-Tunnel splits use seeds
20260715--20260719; Wind-Tunnel and Sachs splits use
20260720--20260724.

\paragraph{Penalty selection.}
Nodewise LassoCV RMS-scales predictors within the training fold, fits an intercept, and minimizes mean validation error over 100 log-spaced penalties from
\(\alpha_{\max}=\|X_c^\top y_c\|_\infty/n\) to \(10^{-3}\alpha_{\max}\).
Its integer-valued CV is deterministic and unshuffled, with cyclic coordinate updates.
The scarce-incident benchmark uses ten inner folds,
\(\texttt{max\_iter}=100{,}000\), and tolerance \(10^{-4}\).
Light uses three incident folds with the same limit; Wind and Sachs fits use
five reference and three incident folds with limit 20,000.
The theory assumes a penalty satisfying its joint-score event; these
training-internal choices evaluate a practical counterpart and do not claim
that cross-validation proves that event.

\paragraph{AI disclosure.}
Generative AI assisted manuscript editing, code review, and experiment
orchestration. It was neither treated as an author nor used as a cited
source; the human authors retain full responsibility.

\clearpage
{\small
\bibliographystyle{plainnat}
\bibliography{main}
}
\end{document}